\documentclass{article}
\usepackage[T1]{fontenc}
\usepackage{listings}
\lstdefinestyle{prompt}{basicstyle=\ttfamily,breaklines=true}
\usepackage{iclr2027_conference,times}

\iclrfinalcopy

\usepackage{url}
\usepackage{natbib}
\usepackage{tikz}
\usepackage{booktabs}
\usepackage{amsmath}
\usepackage{tabularx}
\usepackage{listings}
\usepackage{float}
\usepackage{graphicx}
\usepackage{enumitem}
\usepackage{hyperref}
\usepackage{xcolor}
\usepackage{fvextra}
\DefineVerbatimEnvironment{PromptExcerpt}{Verbatim}{
  fontsize=\scriptsize,
  breaklines=true,
  breakautoindent=true,
  breaksymbolleft=\tiny\ensuremath{\hookrightarrow}
}

\hypersetup{colorlinks=true,linkcolor=blue,citecolor=blue,urlcolor=blue}

\title{Hob-VL: A Benchmark for Visually Grounded Boolean Reasoning}
\author{Yuzhou Wang, Emile Anand, Ijay Narang \\ Georgia Institute of Technology \\ \texttt{\{ywang3694,emile,inarang3\}@gatech.edu}}
\date{}
\begin{document}
\maketitle

\begin{abstract}
Reliable visual reasoning requires composing multiple visual observations and returning consistent answers to logically equivalent questions. We introduce Hob-VL, a benchmark for visually grounded Boolean reasoning. 
Hob-VL comprises two tasks: (1) evaluating whether a Boolean rule holds in an image, and (2) identifying the (unique) object satisfying a Boolean description. Hob-VL contains 6,000 human-verified balanced Yes/No questions, each defined by a Boolean combination of ten visual statements, across 1,000 generated scenes and 46 diverse labeled photographs, along with 1,000 object-identification questions over the same photographs. Our question families are deliberately constructed to challenge reasoning through misleading local cues and nested logical operations, and include symbolic and structured natural-language presentations. Across eight model configurations with thinking disabled or minimized, Boolean accuracy ranges from 48.52\% to 50.57\%, while the identification accuracy reaches at most 43.0\%. 
A thinking-enabled GLM configuration achieves uneven gains while retaining substantial errors and inconsistencies. 
Hob-VL exposes these failures through executable reference answers and matched evaluations.\looseness=-1
\end{abstract}

\section{Introduction}

Understanding how vision-language models (VLMs) interpret and reason about images is a longstanding research goal \citep{antol2015vqa,johnson2017clevr,hudson2019gqa}. Many visual questions require identifying relevant objects, attributes, and relations, then combining these observations according to the question's meaning. Yet answering individual visual questions correctly does not ensure correct answers to their logical combinations \citep{gokhale2020vqalol}, and answers may change across meaning-preserving rephrasings \citep{shah2019cycle}. Reliable visual reasoning therefore requires both accurate composition and consistency across equivalent formulations.

These capabilities matter when visual observations are combined with structured information to answer queries or guide decisions. For instance, EHRXQA links patient records to chest X-ray images \citep{bae2023ehrxqa}, EQUI-VOCAL synthesizes queries for video events involving object attributes and relations \citep{zhang2023equivocal}, and DriveLM connects perception, prediction, and planning through visual question answering \citep{sima2024drivelm}. In an industrial application, a warehouse query might ask for a package that is damaged, is beside a loading bay, is not covered, and is not inside a container \citep{10448946,su14137781}. Another illustrative application is to locate plants that are not under automated irrigation and either (i) have yellow leaves and are wilting or (ii) show pest damage.\citep{zhang2024goodcaptioningbadcounting, rs14225738, soler2024combiningdeeplearningstreet}. Answering these queries correctly requires evaluating the specified combination of conditions for the same object.\looseness=-1

Diagnostic benchmarks have made these capabilities amenable to systematic study. For instance, VQA~v2 pairs similar images requiring different answers to the same question to counter language priors \citep{goyal2017vqa2}, CLEVR uses controlled scenes and executable question programs \citep{johnson2017clevr}, and GQA combines structured question generation with consistency evaluation on photographs \citep{hudson2019gqa}. CLOSURE further shows that strong CLEVR performance need not transfer to new combinations of familiar referring expressions \citep{bahdanau2019closure}. We build on these diagnostics to ask whether VLMs can evaluate
explicit Boolean combinations of visual facts and preserve their
answers when the same function is expressed differently.

Boolean logic provides a natural and precise setting in which to investigate these ideas \citep{lee-etal-2026-fol}. Under this lens, in the formal logic terminology, the \emph{literals} (Boolean variables), stem from directly observable visual statements, while \emph{conjunction} ($\land$, the Boolean `AND'), \emph{disjunction} ($\lor$, the Boolean `OR'), and \emph{negation} ($\neg$, the Boolean `NOT') determine how those values combine to form an answer. Together, this set of operators can express any Boolean function over a finite set of binary facts \citep{odonnell2014}, providing a rich language for composing such conditions. One immediate advantage here is that their explicit semantics allow reference answers to be computed from annotated observations and equivalences between expressions to be checked exactly. 


Within this setting, the demands of composition can be characterized more precisely than by question length alone. For instance, one false input condition might settle a long conjunction, whereas other functions might require several input values to be known. Boolean-function analysis quantifies input influence (how often changing an input changes the answer) and certificate size (the minimum number of fact values sufficient to determine the answer for a given instance) as well as predictability from limited evidence~\citep{odonnell2014}. Nested expressions additionally organize computation into intermediate subproblems, motivating tests of successive composition, and a strong benchmark should record properties of the expression tree, such as depth and connective alternation. Theoretical studies of transformer expressivity establish that intermediate generation can extend computational capabilities under specific architectural assumptions \citep{merrill2024expressive,li2024serial,anand2026continuouslatentcontextsenable,anand2026smatattentionstructuredlongcontextsequence}, motivating our comparison of inference configurations.

Another requirement of a good benchmark is consistency when the same conditions are expressed differently. For instance, by De Morgan's law, $A\land B=\neg(\neg A\lor\neg B)$ and therefore any corresponding prompt must yield the same answer for the same image \citep{demorgan1847formallogic}. Such comparisons fix the visual evidence and Boolean function while changing the expression a model must interpret. Agreement measures consistency, but two agreeing answers can both be incorrect. We therefore report both disagreement and
the fraction of equivalent pairs answered correctly on both members.

We introduce \textbf{Hob-VL}, a benchmark that combines exact Boolean-function annotations with paired evaluations of visually grounded reasoning: equivalent expressions within a presentation format, and symbolic versus structured natural-language presentations of the same question. The benchmark contains $6{,}000$ balanced Yes/No questions over $1{,}000$ generated scenes and $46$ labeled photographs, together with $1{,}000$ object-identification questions over the same photographs. The Yes/No questions combine ten visual statements and form 3,000 human-verified original/De Morgan-rewrite pairs, while the identification questions combine two to five spatial predicates. We make each question available in \emph{both} presentation formats, enabling separate comparisons of equivalent expressions and symbolic versus natural-language presentation. We illustrate these tasks in Figure~\ref{fig:overview}.\looseness=-1

Our evaluation covers eight VLMs in nine configurations, collecting 126K responses. Across the eight configurations with disabled/minimized thinking, Boolean accuracy remains near 50\% (surprisingly, the same accuracy a coin flip between the answers!), while object-identification accuracy reaches at most $43.0\%$ under the recorded scoring policies. The thinking-enabled GLM configuration achieves substantially higher accuracy on symbolic Boolean evaluation and both identification formats, but improves little on natural-language Boolean evaluation and still frequently gives conflicting answers to equivalent expressions. These findings highlight persistent difficulties in visually grounded composition and demonstrate the importance of reporting evaluation metrics such as item accuracy, pairwise correctness, disagreement, and output validity together.\looseness=-1

\begin{center}\begin{figure}[htbp]\begin{center}
\centering
\begin{minipage}[c]{0.27\linewidth}
\centering
\textbf{Image}\par\smallskip
\begin{tikzpicture}[x=0.8cm,y=0.8cm]
 \filldraw[fill=red!75!black,draw=black]
  (0,0) rectangle (0.8,0.8);
 \node at (0.4,-0.35) {\textbf{A}};
 \filldraw[fill=blue!65,draw=black] (2,0.4) circle (0.42);
 \node at (2,-0.35) {\textbf{B}};
\end{tikzpicture}
\end{minipage}\hfill
\begin{minipage}[c]{0.69\linewidth}
\small
\textbf{Yes/No evaluation: answer Yes or No}\par\smallskip
Condition 1: ``A is red and B is not a square.''\par
Condition 2: ``It is not the case that A is not red or B is a square.''\par
Both have answer \textbf{Yes} for this image.\par\medskip
\textbf{Object identification: answer with one label}\par\smallskip
Description: ``The object is left of B and is not above B.''\par
The answer is \textbf{A}, using object centers.
\end{minipage}
\caption{Simplified illustrations of the two  tasks. The
Yes/No example uses logically equivalent Boolean expressions with the same image and answer, while changing the wording. The identification example asks for the label of the object satisfying the description. We provide instances of our released examples in Figure \ref{fig:released-task-examples} and Appendix  \ref{app:release-examples}.}
\label{fig:overview}
\end{center}
\end{figure}
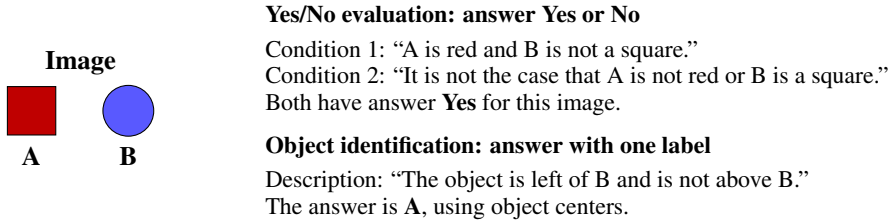
 \end{center}

\section{Related Work}
\label{sec:related}

\textbf{Compositional visual reasoning.}
VQA established large-scale evaluation of answers to natural-language questions about images \citep{antol2015vqa}, and VQA~v2 introduced complementary images to reduce exploitable language priors \citep{goyal2017vqa2}. CLEVR combines generated scenes with executable question programs and measures effective question size through answer-preserving program pruning \citep{johnson2017clevr}. GQA generates compositional questions from scene graphs of photographs and evaluates consistency using entailment relations \citep{hudson2019gqa}. CLOSURE tests CLEVR-trained models on unfamiliar combinations of known referring expressions \citep{bahdanau2019closure}.
Hob-VL adds exhaustive truth-table annotations for ten-input Boolean
functions and pairs each base Yes/No expression with an equivalent
rewrite, and its evaluation measures correctness and consistency on the
released items.

\textbf{Multimodal logical reasoning.}
VQA-LOL directly studies Boolean combinations of visual questions, including conjunction, disjunction, and negation \citep{gokhale2020vqalol}. LogicVista and VisuLogic evaluate diverse forms of reasoning in visual contexts \citep{xiao2024logicvista,xu2025visulogic}. MuSLR more specifically tests multimodal inference grounded in propositional, first-order, and non-monotonic rules \citep{xu2025muslr}, while MM-CondChain constructs visually grounded conditional chains using a programmatically verifiable intermediate representation \citep{shen2026mmcondchain}. Hob-VL focuses on evaluating supplied Boolean functions and selecting objects that satisfy Boolean descriptions. Its Yes/No task supports exhaustive function analysis and verified expression rewrites, and both tasks support matched symbolic and structured natural-language presentations.

\textbf{Equivalent formulations and logical sensitivity.}
VQA-Rephrasings and Global PIQA test robustness to human-written visual-question rephrasings \citep{shah2019cycle,chang2026globalpiqaevaluatingcommonsense}. In text, RobustLR evaluates minimal logical edits and equivalence conditions \citep{sanyal2022robustlr}, while CRTBench includes both equivalence-preserving and answer-reversing reformulations \citep{gu2026controlled}. FACTOR-Bench evaluates logical constraints through dual-encoder scoring and includes an equivalence split with De Morgan transformations, double negation, and commutativity \citep{alshehri2026factor}. Logic Unseen introduces a separate visual LogicBench for logic-aware retrieval and caption selection under logical perturbations \citep{zhou2026logicunseen}. Hob-VL compares generated answers across verified equivalent expressions and, separately, across symbolic and structured natural-language presentations. For equivalent expression pairs, it reports any disagreement among pairs with two valid outputs and the proportion of all pairs answered correctly on both members.

\textbf{Textual logic and Boolean evaluation.}
ProofWriter generates implications and proofs from natural-language rules and facts \citep{tafjord2021proofwriter}, FOLIO pairs natural-language inference with first-order logical annotations \citep{han2024folio}, and the textual LogicBench organizes examples around individual inference rules \citep{parmar2024logicbench}. The Boolean Expressions task in BIG-Bench Hard evaluates expressions containing explicit True/False constants \citep{suzgun2023challenging}. Hob-VL obtains the inputs from visual evidence and characterizes the resulting functions using Boolean-function analysis \citep{odonnell2014}. SATBench asks whether constraints expressed as natural-language puzzles admit a satisfying assignment \citep{wei2025satbench}. 
In Hob-VL's Yes/No task, the image determines assignments, and the model evaluates the supplied expression at that assignment. Once the literals are known, the evaluation runtime is linear in the expression's tree size.\looseness=-1

\textbf{Grounding and interpretation.}
Reasoning performance in language models can vary with premise order \citep{chen2024premise} and added input length \citep{levy2024same}, even when the required inference is unchanged. \cite{nagar2024zeroshot} compare visual embeddings with textual scene descriptions as inputs to underlying language models, illustrating the value of such controls for investigating visual reasoning. These findings motivate distinguishing a question's recorded logical
properties from the processes responsible for model errors. Our paired evaluations measure end-to-end accuracy and consistency across equivalent expressions and symbolic and natural-language presentations.

\section{Benchmark Tasks and Paired Presentations}
\label{sec:release}
\label{sec:properties}

Hob-VL\footnote{\url{https://github.com/Yuzhou-Wang99/Hob-VL}} evaluates two tasks: deciding whether a Boolean condition holds in an image and identifying the object that satisfies a Boolean description. Yes/No evaluation uses generated scenes and photographs; identification uses the same photographs, giving the three tracks in Table~\ref{tab:release}. We count a base expression and its equivalent rewrite as two
questions, each with a distinct identifier. Each question has two
presentation formats, yielding 7,000 questions and 14,000 prompts.
The benchmark has no training split. Each request contains a labeled image and one prompt; reference answers, annotated truth values, and construction metadata are used only for validation and scoring.

\begin{table}[htbp]
\centering\small
\begin{tabular}{@{}lrrrr@{}}
\toprule
Track & Images & Questions & Yes / No & Prompts \\
\midrule
Yes/No, generated scenes & 1{,}000 & 4{,}000 & 2{,}000 / 2{,}000 & 8{,}000 \\
Yes/No, photographs & 46 & 2{,}000 & 1{,}000 / 1{,}000 & 4{,}000 \\
Identification, photographs & 46 & 1{,}000 & --- & 2{,}000 \\
\midrule
Total & 1{,}046 & 7{,}000 & --- & 14{,}000 \\
\bottomrule
\end{tabular}
\caption{Hob-VL counts. Each question has one symbolic and one structured natural-language prompt. Yes/No counts include base expressions and equivalent rewrites, and each Yes/No track has equal Yes and No counts.\\}
\label{tab:release}
\end{table}
\textbf{Yes/No evaluation.}
\label{sec:release-yesno}
Each question combines ten visual statements, such as ``Object A is blue,'' using AND, inclusive OR, and NOT. Statement $i$ corresponds to a variable $x_i$, whose truth value $a_i$ is determined by the image. The expression represents a function $F:\{0,1\}^{10}\to\{0,1\}$, whose correct answer is \texttt{Yes} if $F(a)=1$ and \texttt{No} otherwise. While Variables may repeat, always referring to the same statement, every variable is essential: for each $i$, some assignment changes the output when only $x_i$ is flipped. Nevertheless, fewer than ten facts can suffice for a particular depicted instance (Section~\ref{sec:dependence}).

\textbf{Object identification.}
\label{sec:release-identification}
Each predicate is evaluated separately for each candidate object. For instance, ``the candidate is left of B'' AND NOT ``the candidate is above C'' selects objects satisfying both spatial requirements. Released descriptions combine two to five spatial predicates, and the model must return the single satisfying object's label. In Hob-VL, all listed objects are candidates, including objects named as
references in the description; moreover, directions refer to designated object
centers in image coordinates, and self-relations such as ``A is left
of A'' are false. In particular, we maintain the invariant that an object is not left, right, above, or below itself. We elaborate in the Appendix on how our construction handles
approximate reference points and uncertain comparisons.

\begin{figure}[hbt!]
\centering
\begin{minipage}{\linewidth}
\small

\textbf{(a) Yes/No evaluation: an excerpt from a released control question}
\par\smallskip
\begin{minipage}[t]{0.35\linewidth}
\vspace{0pt}
\centering
\includegraphics[width=\linewidth]{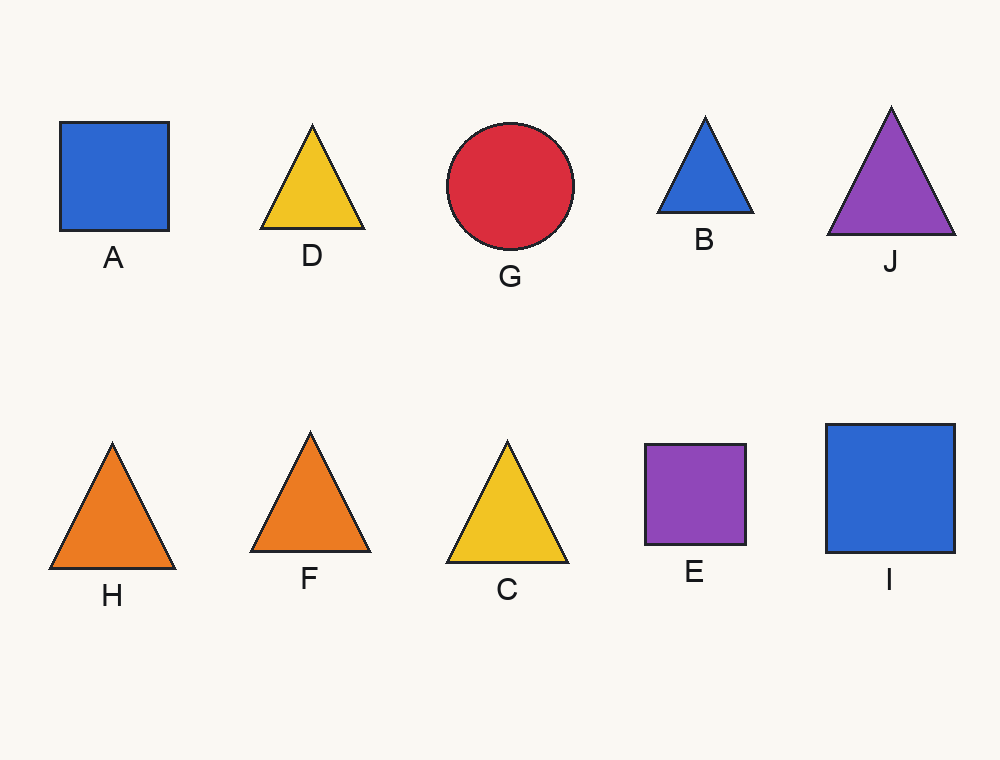}
\end{minipage}\hfill
\begin{minipage}[t]{0.62\linewidth}
\vspace{0pt}
\raggedright
\textbf{Selected statements and symbolic clause}\par
$p_6$: Object D is orange.\par
$p_7$: Object G is a triangle.\par
$p_8$: Object J is a circle.\par\smallskip
\texttt{(p6 OR (p7 OR p8))}\par\smallskip
\textbf{Corresponding NL clause}\par
\begin{itemize}[label={$-$},leftmargin=1em,nosep,topsep=0pt]
\item At least one of the following conditions holds, possibly both:
  \begin{itemize}[label={$-$},leftmargin=1em,nosep,topsep=0pt]
  \item Object D is orange.
  \item At least one of the following conditions holds, possibly both:
    \begin{itemize}[label={$-$},leftmargin=1em,nosep,topsep=0pt]
    \item Object G is a triangle.
    \item Object J is a circle.
    \end{itemize}
  \end{itemize}
\end{itemize}
\end{minipage}

\par\smallskip
\noindent\resizebox{\linewidth}{!}{%
\textbf{Reference answer: No.} D is yellow, G is a circle, and J is a triangle. The displayed clause is false, so the full conjunction is false.%
}

\par\medskip\hrule\medskip

\textbf{(b) Object identification: a complete released condition}
\par\smallskip
\begin{minipage}[t]{0.29\linewidth}
\vspace{0pt}
\centering
\includegraphics[width=\linewidth]{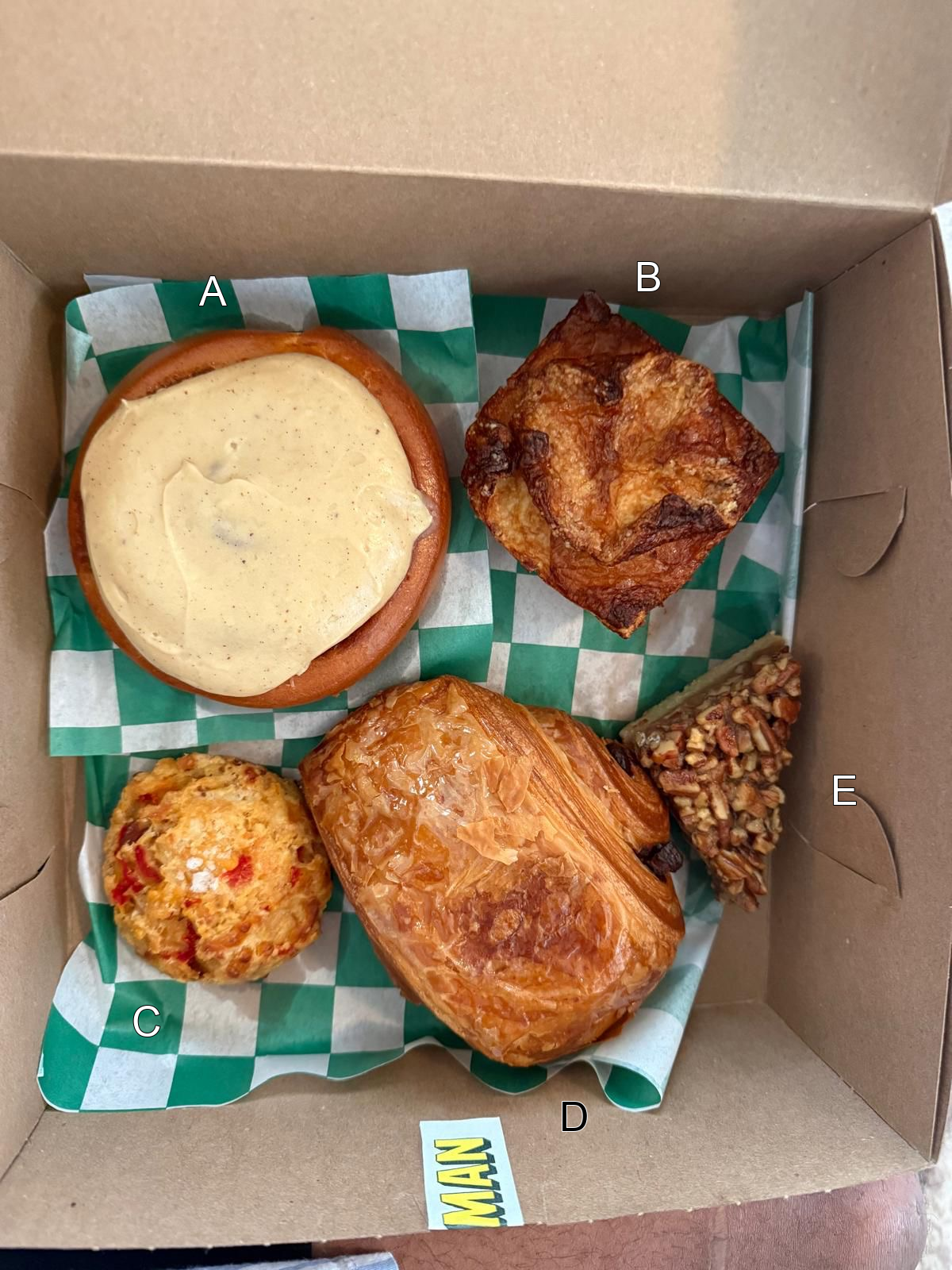}
\end{minipage}\hfill
\begin{minipage}[t]{0.68\linewidth}
\vspace{0pt}
\raggedright
\textbf{Candidates: A, B, C, D, E}\par
$p_1$: The candidate object is to the right of object B.\par
$p_2$: The candidate object is above object C.\par
$p_3$: The candidate object is to the left of object D.\par
$p_4$: The candidate object is above object E.\par\smallskip
\textbf{Symbolic condition}\quad
$(p_1\lor\neg p_2)\land(p_3\lor p_4)$\par\smallskip
\textbf{Corresponding NL condition}\par
\begin{itemize}[label={$-$},leftmargin=1em,nosep,topsep=0pt]
\item Both of the following conditions hold:
  \begin{itemize}[label={$-$},leftmargin=1em,nosep,topsep=0pt]
  \item At least one of the following conditions holds, possibly both:
    \begin{itemize}[label={$-$},leftmargin=1em,nosep,topsep=0pt]
    \item The candidate object is to the right of object B.
    \item It is not the case that the following condition holds:
      \begin{itemize}[label={$-$},leftmargin=1em,nosep,topsep=0pt]
      \item The candidate object is above object C.
      \end{itemize}
    \end{itemize}
  \item At least one of the following conditions holds, possibly both:
    \begin{itemize}[label={$-$},leftmargin=1em,nosep,topsep=0pt]
    \item The candidate object is to the left of object D.
    \item The candidate object is above object E.
    \end{itemize}
  \end{itemize}
\end{itemize}
\end{minipage}

\par\smallskip
\noindent\resizebox{\linewidth}{!}{%
\textbf{Reference answer: C.} C is the only candidate satisfying the complete condition. Objects used as references remain candidates.%
}
\par

\end{minipage}
\caption{Released Hob-VL examples. (a) One of eight clauses in a ten-statement sparse conjunctive normal form (CNF) question.
(b) A complete object-identification condition. Instructions are omitted and notation adapted for readability. Reference answers are shown only for the reader. Appendix~\ref{app:release-examples} provides exact prompts}
\label{fig:released-task-examples}
\end{figure}

\textbf{Comparisons.}
First, each question has a \emph{symbolic} and a \emph{structured natural-language} (NL) presentation, sharing its identifier, image, and reference answer. The symbolic prompt defines the visual
statements and gives a fully parenthesized expression. The natural language (NL) prompt renders the same abstract syntax tree (AST) as nested conditions
with explicit scope. Second, each base Yes/No expression $F$ has a De Morgan rewrite $\widetilde F$ that changes its syntax while preserving its truth value on every assignment. The base and rewritten questions have distinct identifiers linked by an expression-pair identifier. Thus, each pair has four prompts:
\begin{table}[htbp]
\centering
\begin{tabular}{c|cc}
 & Symbolic & Natural language \\ \hline
Base question & $F$ & $\mathrm{NL}(F)$ \\
Rewritten question & $\widetilde{F}$ & $\mathrm{NL}(\widetilde{F})$
\end{tabular}
\caption{Base/De Morgan-rewritten expressions in symbolic vs structured natural-language formats.}
\label{tab:prompt-formats}
\end{table}
Comparing columns measures sensitivity to presentation format, and comparing rows measures consistency across equivalent expressions within a format. Hob-VL has  3K base-rewrite pairs, which forms 6K Yes/No questions and 3K equivalent-expression pairs in \emph{each} format. The identification tasks add 1K questions in both formats without a separate rewrite collection. So, a full evaluation makes 14K separate API requests per model configuration, where the underlying questions share images and expressions.

For example, let $A$ denote ``Object A is red'' and $B$ denote
``Object B is a square.'' The condition $A\land\neg B$ can be
rewritten as $\neg(\neg A\lor\neg\neg B)$ by applying De Morgan's
law once. If $A=1$ and $B=0$, both expressions are true, and exhaustive
comparisons verify that they agree for every assignment. We keep
the double negation in this example to show the three NOT nodes
introduced by the rewrite. This construction is applied to
various subexpressions of the ten-input benchmark expressions.

\section{Construction and Validation}
\label{sec:construction}

We construct Yes/No questions in four stages: annotate visual statements, sample Boolean expressions, compute solutions and diagnostics, and generate equivalent expressions. The annotations determine the truth values of the ten visual statements, and a Boolean interpreter evaluates each expression under resulting assignment.  In turn, object-identification questions use the same photographs but a separate procedure that evaluates the description for every candidate object. We describe these procedures and their validation below. \looseness=-1

\subsection{Visual facts and expression families}
\label{sec:visual-facts}

Each synthetic generated scene contains ten labeled objects, drawn from three shapes (circle, square, and triangle) and six colors. The ten statements each query one color or shape of a different object. Therefore, all $2^{10}$ truth assignments are realizable across possible generated scenes. Photographic statements describe visible attributes and spatial relations recorded in the annotations. 

In each track, let $S$ be the structured scene annotations and let $g_i(S)\in \{0,1\}$ be the annotated truth value of statement $i$. The ten predicates then define the assignment $a(S) = \bigl(g_1(S),\dots,g_{10}(S)\bigr)$, and the reference answer is $F(a(S))$. The model receives the labeled image, statement definitions, and logical condition, but not the annotation record $S$ or the annotated truth values $a(S)$.

\textbf{Expression families.} The families in Table~\ref{tab:families} vary dependence on multiple inputs, conditional selection, and overlapping constraints in sub-expressions. We use XOR ($\oplus$) to mean that an odd number of its inputs are true; a majority ($\operatorname{Maj}_k$) returns true when more than $k/2$ inputs are true. A multiplexer, $M(s,u,v)=(s\land u)\lor(\neg s\land v)$, returns $u$ when $s=1$ and $v$ when $s=0$, where reusing a selector across branches couples their outputs. A literal is a variable or its negation, and a conjunctive normal form (CNF) is an AND of clauses, each an OR of literals. These construction abbreviations are expanded into binary AND/OR and unary NOT before prompting or size measurement.\looseness=-1

 \begin{table}[hbt!]
\centering
\footnotesize
\setlength{\tabcolsep}{3pt}
\renewcommand{\arraystretch}{1.0}
\begin{tabularx}{\linewidth}{@{}
  >{\raggedright\arraybackslash}p{0.21\linewidth}
  >{\raggedright\arraybackslash}p{0.46\linewidth}
  >{\raggedright\arraybackslash}X@{}}
\toprule
Family ($n$) & Operation and example & Diagnostic role \\
\midrule
Parity of conjunctions\newline ($1{,}000$)
& Parity of five AND-pair outputs:
  $(1,1,0,0,0)$ gives false;
  $(1,1,1,0,0)$ gives true.
& Cross-fact dependence; combining local results rather than
  counting true statements. \\
\addlinespace[2pt]
Parity of majorities\newline ($1{,}000$)
& XOR two five-input majorities. Groups with three and one true
  inputs yield different majorities, hence true.
& Composing threshold decisions while preserving fact grouping. \\
\addlinespace[2pt]
Nested multiplexers\newline ($1{,}500$)
& Select branches via $M(1,u,v)=u$ and $M(0,u,v)=v$;
  nest selections and reuse inputs.
& Nested evaluation; consistent reuse of facts across branches. \\
\addlinespace[2pt]
Dense CNF\newline ($1{,}000$)
& Conjoin 24 five-literal clauses. Any clause with five false
  literals makes the formula false.
& Overlapping constraints coupled through shared variables. \\
\addlinespace[2pt]
Sparse chain CNF\newline ($500$)
& Conjoin eight three-literal clauses; adjacent clauses share
  two variables. Any false clause makes the formula false.
& Local-interaction control against dense CNF;
  short false-answer certificates. \\
\addlinespace[2pt]
Polarity-trap composition\newline ($500$)
& Core $D$ forces a signed input false despite three unnegated
  and two negated mentions. Compose with parity $P$: $D\oplus P$.
& Occurrence-majority cue favors the wrong value
  for satisfying the core. \\
\addlinespace[2pt]
Ten-input parity\newline ($500$)
& Odd parity of ten signed inputs: three true inputs give true;
  four give false.
& Broad-dependence control with a simple counting procedure. \\
\bottomrule
\end{tabularx}
\caption{Construction families, evaluation examples, and diagnostic
roles. Question counts include base and De Morgan variants.
Examples precede optional output negation; signed inputs are
variables or their negations. Appendix \ref{app:family-construction}
provides all templates, sampling procedures, and evaluations.}
\label{tab:families}
\end{table}

\textbf{Selection and answer balance.}
The sampler randomizes variable order and literal signs, retains functions with true-output fraction $q(F)\in[0.1,0.9]$, and requires every input's influence to be at least $0.02$ (defined below). It negates the complete expression when necessary to meet the assigned Yes/No quota, then computes the reference answer from the recorded image facts. This produces balanced labels within each synthetic scene and across each Boolean track. The resulting functions follow this selected construction distribution.

\subsection{Measuring dependence beyond length}
\label{sec:dependence}

For each $F$, enumerate all $2^{10}=1{,}024$ assignments. Under the uniform distribution on $\{0,1\}^{10}$, let
\begin{equation}
 q(F):=\Pr_x[F(x)=1],\qquad
 \operatorname{Inf}_i(F):=\Pr_x[F(x)\ne F(x\oplus e_i)],
 \label{eq:influence}
\end{equation}
where $x\oplus e_i$ flips bit $i$ \citep{odonnell2014}. Positive influence means that a variable is essential: changing it affects the output on at least one assignment, but it may not mean that every instance requires observing that input.  For the depicted assignment $a$, the minimum certificate size is
\begin{equation}
 c_F(a)=\min_{T\subseteq\{1,\ldots,10\}}
 \bigl\{|T|:\ \forall z \in \{0,1\}^{10},\;z_T=a_T\Longrightarrow F(z)=F(a)\bigr\}.
 \label{eq:certificate}
\end{equation}
A \emph{certificate} is a subset of observed values sufficient to fix the answer regardless of the unobserved values. We use its size as a proxy to measure the sufficient information needed to answer a query. We also measure the maximum prediction accuracy obtainable from any fixed pair of inputs over the uniform assignment distribution (Appendix~\ref{app:construction}), since a large certificate need not preclude good approximate prediction from fewer facts. Among the 6K Yes/No questions, 4,166 satisfy our criterion that $\min_i\operatorname{Inf}_i(F)\ge0.1$ and $c_F(a)\ge5$. For 3,030 questions, no predictor based on a fixed pair of inputs exceeds 60\% accuracy. Certificate sizes range from three to ten, despite every function having ten essential inputs. These counts include both members of each base--rewrite pair. Appendix~\ref{app:construction} discusses the structural distributions, certificate bounds, and the treatment of realizable visual assignments.

\subsection{Rewrites, language, and identification}
\label{sec:release-language}

We construct a paired variant by applying De Morgan's identities to selected subexpressions:
\[
  A\land B\equiv\neg(\neg A\lor\neg B),
  \qquad
  A\lor B\equiv\neg(\neg A\land\neg B).
\]
Each application swaps one AND/OR operator and adds three NOT nodes to the abstract syntax tree (AST). We apply at most three disjoint rewrites, with a total increase of at most 15\% of the base expression's node count, and exhaustively verify equivalence for all 3,000 base--rewrite pairs.

\textbf{Prompt generation.}
Each expression is rendered symbolically and with a fixed recursive natural-language grammar. AND requires both child conditions; OR requires at least one, including the case in which both hold; and NOT negates its entire child condition. In the natural-language rendering, each occurrence of a variable is replaced by its full visual statement. The natural-language prompts therefore preserve the supplied tree and explicitly mark logical scope, and are deterministic renderings.  We specify the grammatical detail in Appendix~\ref{app:language}.

\textbf{Object-identification construction.}
Nine templates combine two to five spatial predicates, evaluated using the annotated object reference points. For distinct objects, a horizontal comparison is unknown when their reference points differ by at most 6\% of image width, and the corresponding threshold for a vertical comparison is 6\% of image height. As previously mentioned, all self-relations are false. Therefore, we retain an item only if all predicates for its target have known values and the target is the unique satisfying candidate under every completion of the unknown values. Thus the target must satisfy the description, and every other candidate must fail it, regardless of how unknown comparisons are resolved. We require at least two known predicate values to distinguish the target from the other candidates. Separately, we fix self-relations to false and treat all relations between distinct objects as unknown, retaining a question only if at least two candidates can still satisfy the description. Finally, within each image, we duplicate descriptions that are Boolean-equivalent over the same named predicates.

\subsection{Validation and dataset coverage}
\label{sec:quality}

Yes/No validation uses an independent parser and a NumPy evaluator to replay rewrites and recompute logical and structural annotations. The validation suite reports 681,691 checks and evaluates 6,144,000 truth-table rows, covering all 1,024 assignments for each of the 6,000 expressions. A separate natural-language parser reconstructs all 6,000 Yes/No ASTs and verifies their reference answers. Automated geometric and sampled-pixel checks cover every generated image. Identification validation performs all 7,524 candidate evaluations and confirms that each recorded target is uniquely selected under the specified unknown-value semantics. These procedures verify logical consistency conditional on the recorded visual facts. Finally, we augment the quality assurance of these facts through a comprehensive human annotation review.

The 3,000 base expressions represent 2,746 distinct truth tables. Displayed expressions contain 54--370 AST nodes and have edge depth 6--15. Each generated image supports four Yes/No questions. The photograph tracks share 46 images from 45 scene families, and the identification questions use 241 distinct photograph--target combinations. So, the questions share visual evidence, while some also share a target or Boolean function.  The median symbolic Yes/No prompt contains 371 whitespace-delimited words, compared with 1,911.5 for its natural-language counterpart. Format comparisons also compare prompts of different lengths, and we group records by the expression's root connective and predict each group's majority answer, using the same records to determine and evaluate that prediction. This achieves 52.5\% accuracy, indicating residual association between surface form and the answer. Appendices~\ref{app:construction} and~\ref{app:checks} provide further coverage statistics and validation commands.

\section{Experiments}
\label{sec:experiments}
\label{sec:validation}

\subsection{Evaluation protocol}

We evaluate eight VLMs in nine configurations (Table~\ref{tab:model-accuracy}).  Seven configurations use disabled-thinking settings or non-thinking Instruct variants, and Gemini uses its minimal-thinking setting. We refer to these eight configurations as the \emph{Thinking disabled or minimized}, without asserting an absence of all internal reasoning. We also evaluate GLM-4.6V-FlashX with native thinking enabled. Each configuration is evaluated on both tasks in both presentation formats, yielding 126K final responses across the nine configurations. Appendix~\ref{app:experiments} records the requested API model identifiers and inference settings.

Each request contains an image and a prompt, without reference answers or access to search, solvers, or code execution, with default temperatures. Prepared images have a maximum edge length of $2{,}048$ pixels, though providers may apply additional preprocessing. The output-token cap is $2{,}048$ for the disabled/minimal group and $4{,}096$ for thinking-enabled GLM. While some answer-format instructions differ across models, we ensure that within each configuration, the symbolic and natural-language conditions use the same inference settings and answer-format policy.

\textbf{Scoring.} We parse the complete final response as either \texttt{Yes}/\texttt{No} or a single candidate label. Invalid, missing, and incomplete responses count as incorrect, and we do not extract answers from explanations. Appendix \ref{app:experiments} elaborates on our permitted case, whitespace, punctuation, and markdown. Requests that failed due to API errors are retried with the same settings, whereas completed incorrect or invalid responses are retained. Every item in our evaluation has a final response after retries. \looseness=-1

\textbf{Metrics.} In each format, accuracy is computed over all 6K Boolean questions or all 1K identification questions, including invalid responses. For each Boolean format, \emph{both-correct accuracy} is the fraction of all 3K equivalent-expression pairs for which both answers are correct. The \emph{conflict rate} is the fraction of pairs with two valid outputs whose answers disagree. We report descriptive results without independent-trial confidence intervals because questions share scenes and expressions.

 \begin{table}[htbp]
\centering
\footnotesize
\setlength{\tabcolsep}{4pt}
\renewcommand{\arraystretch}{1.0} 
\begin{tabular*}{\linewidth}{@{\extracolsep{\fill}}lrrrr@{}}
\toprule
& \multicolumn{2}{c}{Boolean evaluation} & \multicolumn{2}{c}{Object identification} \\
\cmidrule(lr){2-3}\cmidrule(l){4-5}
Configuration & Symbolic & Natural Language & Symbolic & Natural Language \\
\midrule
\multicolumn{5}{l}{\emph{Thinking disabled or minimized}} \\
GPT-5.6 Luna              & 49.05 & 48.67 & \textbf{35.70} & 40.00 \\
Claude Haiku 4.5          & 49.37 & 50.22 & 32.20 & \textbf{43.00} \\
Grok 4.3                 & 48.75 & 49.57 & 21.60 & 25.70 \\
Gemini 3.5 Flash-Lite     & 48.60 & 48.52 & 27.10 & 41.10 \\
DeepSeek Flash           & 49.62 & 49.67 & 16.50 & 24.30 \\
GLM-4.6V-FlashX (off)     & 48.58 & \textbf{50.37} & 31.30 & 35.00 \\
Llama 4 Scout            & 49.98 & 48.93 & 27.80 & 33.30 \\
Qwen3-VL-8B-Instruct      & \textbf{50.57} & \textbf{50.37} & 23.00 & 28.60 \\
\midrule
\multicolumn{5}{l}{\emph{Thinking enabled}} \\
GLM-4.6V-FlashX (on)      & 63.50 & 51.20 & 73.70 & 72.10 \\
\bottomrule
\end{tabular*}
\caption{Accuracy (\%) for each configuration and presentation format. Each Boolean column contains 6K questions, and each identification column contains 1K. Invalid responses count as incorrect. Bold indicates the largest observed score in the group, including ties. Thinking-enabled GLM uses a $4{,}096$-token output cap, and other configurations use $2{,}048$ tokens. Appendix~\ref{app:experiments} documents the configurations, model identifiers, prompt policies, and scoring.}
\label{tab:model-accuracy}
\end{table}

\subsection{Accuracy across tasks and formats}

\textbf{Boolean accuracy stays close to the constant-label baseline.}
Across the disabled/minimal configurations and both formats, Boolean accuracy ranges from $48.52\%$ to $50.57\%$, close to the $50\%$ constant-label baseline, with no consistent advantage for either format. 

\textbf{Identification accuracy is higher in natural language.}
For identification, uniform guessing among each question's listed candidates has expected accuracy of $16.90\%$. Always predicting B, which appears in every candidate list, achieves $19.20\%$; this descriptive baseline selects the most frequent reference label using the evaluation labels.
Identification accuracy is higher in natural language for all eight configurations, with gains of $3.7$--$14.0$ percentage points. Symbolic accuracy ranges from $16.5\%$ to $35.7\%$, compared with $24.3\%$ to $43.0\%$ in natural language. GPT-5.6 Luna, GLM with thinking disabled, and Llama achieve natural-language identification accuracies that exceed their symbolic accuracies by 4.3, 3.7, and 5.5 percentage points, respectively.

\textbf{Thinking-enabled GLM shows substantial but uneven gains.}
Its accuracy rises from $48.58\%$ to $63.50\%$ on symbolic Boolean questions, from $31.30\%$ to $73.70\%$ on symbolic identification,
and from $35.00\%$ to $72.10\%$ on Natural Language identification. Natural Language Boolean accuracy changes only from $50.37\%$ to $51.20\%$. The symbolic Boolean gain appears for both generated scenes and photographs (Appendix~\ref{app:experiments}).\looseness=-1

\subsection{Performance by logical properties}
\label{sec:property-results}

We use the recorded certificate sizes, two-fact predictability, and input
influences to examine performance beyond aggregate accuracy and family labels.
Across the nine configurations, the clearest pooled differences appear for
thinking-enabled GLM (Appendix~\ref{app:logical-property-results}).
Table~\ref{tab:property-certificate} shows that, relative to the thinking-disabled GLM configuration, the corresponding symbolic-accuracy gains are $36.91$, $8.33$,
and $0.22$ percentage points. In this pooled analysis, the observed gains concentrate in questions with smaller instance certificate sizes.

\begin{table}[htbp]
\centering
\footnotesize
\setlength{\tabcolsep}{4pt}
\renewcommand{\arraystretch}{1.0} 
\begin{tabular}{@{}lrrrrr@{}}
\toprule
Certificate & $N$ & Sym. off & Sym. on & NL off & NL on \\
\midrule
3--4 & 1,772 & 45.71 & 82.62 & 52.88 & 62.19 \\
5--6 & 2,856 & 49.02 & 57.35 & 48.98 & 47.41 \\
7--10 & 1,372 & 51.38 & 51.60 & 50.00 & 44.90 \\
\bottomrule
\end{tabular}
\caption{GLM-4.6V-FlashX accuracy (\%) by instance certificate size.
$N$ counts questions per format, including both equivalent-expression variants.
Invalid outputs count as incorrect. Off/on denote thinking-disabled/enabled
configurations with output-token caps of 2,048/4,096, respectively.
Each group contains a different mixture of construction families.}
\label{tab:property-certificate}
\end{table}

Higher two-fact predictability and lower total influence also accompany higher pooled symbolic accuracy for thinking-enabled GLM. Both properties vary within only the two CNF families, and within-family certificate results include reversals. 

\subsection{Consistency and output failures}

\textbf{Higher item accuracy can coexist with conflicting answers.}
With thinking enabled, GLM answers both members correctly in $1{,}354/3{,}000$ symbolic pairs ($45.13\%$), compared with $39.13\%$ with thinking disabled. Despite this increase, both-correct accuracy remains below the $50\%$ constant-label baseline. The conflict rate also increases, from $18.90\%$ to $36.47\%$ among pairs with two valid outputs. Conversely, low conflict can coexist with strong label bias: in natural-language Boolean evaluation, GLM with thinking disabled predicts Yes on $98.60\%$ of all questions, with a conflict rate of only $1.67\%$ and accuracy of $50.37\%$.  \looseness=-1

\textbf{Invalid responses further reduce measured accuracy.}
Thinking-enabled GLM produces $827$ invalid responses, of which $826$ terminate at the token limit, including $536$ in natural-language Boolean evaluation. DeepSeek produces $639$ invalid responses, including $81$ token-limit failures, and Gemini produces $184$, including $10$ token-limit failures. The remaining invalid responses from DeepSeek and Gemini finish generation but fail the final-answer parser. Appendix~\ref{app:experiments} reports these outcomes without extracting answers from explanatory prose. The reported accuracy thus reflects both correctness and the ability to return the requested answer within the specified output budget.\looseness=-1

\section{Conclusion}

Hob-VL combines executable Boolean annotations, equivalent-expression pairs, and matched symbolic and natural-language prompts to evaluate visually grounded composition. Across eight configurations with thinking disabled/minimized, Boolean accuracy is close to the $50\%$ constant-label baseline, while identification accuracy varies substantially across formats. The thinking-enabled GLM configuration achieves higher accuracy in several conditions but still produces frequent conflicts across equivalent expressions. These findings motivate reporting accuracy together with pairwise correctness, conflict rates, label frequencies, and output validity.

\section*{Acknowledgements}
We thank Jan van den Brand, Sarah Liaw, Albert Weng, Aditya Lonkar, and Siddharth Meenachi Sundaram for insightful discussions on this work. Emile Anand is supported by NSF Grant CCF 2338816. Yuzhou Wang and Ijay Narang are supported by NSF grant CCF 2309708.

\section*{Reproducibility Statement}

The construction procedures, validation checks, and prompt grammar are described
in Section~\ref{sec:construction} and Appendices~\ref{app:construction}, \ref{app:language}, and~\ref{app:checks}. Appendix~\ref{app:experiments} documents the requested model
identifiers, inference settings, scoring policies, and procedure for regenerating
the result tables. Saved prompts and prepared images are identified by hashes,
and append-only evaluation histories retain all retries. The reported evaluation
results can be recomputed from the saved histories and reference annotations,
using the final response after retries for each question, presentation format,
and model configuration, and therefore regenerating the tables does not require further model API calls.

\section*{Ethics Statement} 
 Hob-VL is intended to support research on visually grounded Boolean composition. Its results should be interpreted within the scope of its two tasks and image collection. The 46 photographs were taken by the authors. We reviewed the images and prompts for identifying information and sensitive content.

\section*{AI Use Statement}

We used generative AI tools, including OpenAI's Codex and Cognition's Devin, to assist with evaluation and analysis code, preparation of result tables and figures, and manuscript editing. AI assistance also included visual review of photographic facts and object reference-point annotations, as described in
Appendix~\ref{app:image-review}. Human reviewers verified the final dataset annotations. The authors take responsibility for
the final text, code, annotations, and conclusions.

\bibliographystyle{plainnat}
\bibliography{iclr2027_conference}

@inproceedings{alshehri2026factor,
  author = {Alshehri, Sultan and Yang, Zhantao and Zhang, Han and Savvides, Marios},
  title = {Similarity Is Not Logic: Factored Inference for Dual-Encoder Vision-Language Models},
  booktitle = {International Conference on Machine Learning},
  year = {2026},
  url = {https://arxiv.org/abs/2607.23052}
}

@inproceedings{antol2015vqa,
  author = {Antol, Stanislaw and Agrawal, Aishwarya and Lu, Jiasen
    and Mitchell, Margaret and Batra, Dhruv and Zitnick, C. Lawrence
    and Parikh, Devi},
  title = {{VQA}: Visual Question Answering},
  booktitle = {Proceedings of the IEEE International Conference on Computer Vision},
  pages = {2425--2433},
  year = {2015},
  url = {https://openaccess.thecvf.com/content_iccv_2015/html/Antol_VQA_Visual_Question_ICCV_2015_paper.html}
}

@misc{bahdanau2019closure,
  author = {Bahdanau, Dzmitry and de Vries, Harm and O'Donnell, Timothy J.
    and Murty, Shikhar and Beaudoin, Philippe and Bengio, Yoshua
    and Courville, Aaron},
  title = {{CLOSURE}: Assessing Systematic Generalization of {CLEVR} Models},
  year = {2019},
  note = {arXiv:1912.05783; revised 2020},
  url = {https://arxiv.org/abs/1912.05783}
}

@inproceedings{chen2024premise,
  author = {Chen, Xinyun and Chi, Ryan Andrew and Wang, Xuezhi and Zhou, Denny},
  title = {Premise Order Matters in Reasoning with Large Language Models},
  booktitle = {Proceedings of the 41st International Conference on Machine Learning},
  series = {Proceedings of Machine Learning Research},
  volume = {235},
  pages = {6596--6620},
  publisher = {PMLR},
  year = {2024},
  url = {https://proceedings.mlr.press/v235/chen24i.html}
}

@inproceedings{goyal2017vqa2,
  author = {Goyal, Yash and Khot, Tejas and Summers-Stay, Douglas
    and Batra, Dhruv and Parikh, Devi},
  title = {Making the {V} in {VQA} Matter: Elevating the Role of Image Understanding in Visual Question Answering},
  booktitle = {Proceedings of the IEEE Conference on Computer Vision and Pattern Recognition},
  pages = {6904--6913},
  year = {2017},
  url = {https://arxiv.org/abs/1612.00837}
}

@misc{gu2026controlled,
  author = {Gu, Alexander and Chen, Alan},
  title = {Controlled Reformulation Testing for Logical Consistency in Large Language Models},
  year = {2026},
  eprint = {2607.14528},
  archivePrefix = {arXiv},
  primaryClass = {cs.CL},
  url = {https://arxiv.org/abs/2607.14528}
}

@inproceedings{han2024folio,
  author = {Han, Simeng and Schoelkopf, Hailey and Zhao, Yilun
    and Qi, Zhenting and Riddell, Martin and Zhou, Wenfei
    and Coady, James and Peng, David and Qiao, Yujie
    and Benson, Luke and Sun, Lucy and Wardle-Solano, Alexander
    and Szab{\'o}, Hannah and Zubova, Ekaterina and Burtell, Matthew
    and Fan, Jonathan and Liu, Yixin and Wong, Brian
    and Sailor, Malcolm and Ni, Ansong and Nan, Linyong
    and Kasai, Jungo and Yu, Tao and Zhang, Rui
    and Fabbri, Alexander and Kryscinski, Wojciech Maciej
    and Yavuz, Semih and Liu, Ye and Lin, Xi Victoria
    and Joty, Shafiq and Zhou, Yingbo and Xiong, Caiming
    and Ying, Rex and Cohan, Arman and Radev, Dragomir},
  title = {{FOLIO}: Natural Language Reasoning with First-Order Logic},
  booktitle = {Proceedings of the 2024 Conference on Empirical Methods in Natural Language Processing},
  pages = {22017--22031},
  publisher = {Association for Computational Linguistics},
  year = {2024},
  doi = {10.18653/v1/2024.emnlp-main.1229},
  url = {https://aclanthology.org/2024.emnlp-main.1229/}
}

@inproceedings{levy2024same,
  author = {Levy, Mosh and Jacoby, Alon and Goldberg, Yoav},
  title = {Same Task, More Tokens: the Impact of Input Length on the Reasoning Performance of Large Language Models},
  booktitle = {Proceedings of the 62nd Annual Meeting of the Association for Computational Linguistics (Volume 1: Long Papers)},
  pages = {15339--15353},
  publisher = {Association for Computational Linguistics},
  year = {2024},
  doi = {10.18653/v1/2024.acl-long.818},
  url = {https://aclanthology.org/2024.acl-long.818/}
}

@inproceedings{nagar2024zeroshot,
  author = {Nagar, Aishik and Jaiswal, Shantanu and Tan, Cheston},
  title = {Zero-Shot Visual Reasoning by Vision-Language Models: Benchmarking and Analysis},
  booktitle = {2024 International Joint Conference on Neural Networks (IJCNN)},
  pages = {1--8},
  year = {2024},
  doi = {10.1109/IJCNN60899.2024.10650020},
  url = {https://arxiv.org/abs/2409.00106}
}

@inproceedings{parmar2024logicbench,
  author = {Parmar, Mihir and Patel, Nisarg and Varshney, Neeraj
    and Nakamura, Mutsumi and Luo, Man and Mashetty, Santosh
    and Mitra, Arindam and Baral, Chitta},
  title = {{LogicBench}: Towards Systematic Evaluation of Logical Reasoning Ability of Large Language Models},
  booktitle = {Proceedings of the 62nd Annual Meeting of the Association for Computational Linguistics (Volume 1: Long Papers)},
  pages = {13679--13707},
  publisher = {Association for Computational Linguistics},
  year = {2024},
  doi = {10.18653/v1/2024.acl-long.739},
  url = {https://aclanthology.org/2024.acl-long.739/}
}

@inproceedings{sanyal2022robustlr,
  author = {Sanyal, Soumya and Liao, Zeyi and Ren, Xiang},
  title = {{RobustLR}: A Diagnostic Benchmark for Evaluating Logical Robustness of Deductive Reasoners},
  booktitle = {Proceedings of the 2022 Conference on Empirical Methods in Natural Language Processing},
  pages = {9614--9631},
  publisher = {Association for Computational Linguistics},
  year = {2022},
  doi = {10.18653/v1/2022.emnlp-main.653},
  url = {https://aclanthology.org/2022.emnlp-main.653/}
}

@article{shen2026mmcondchain,
  author = {Shen, Haozhan and Yan, Shilin and Xue, Hongwei
    and Lu, Shuaiqi and Tang, Xiaojun and Zhang, Guannan
    and Zhao, Tiancheng and Yin, Jianwei},
  title = {{MM-CondChain}: A Programmatically Verified Benchmark for Visually Grounded Deep Compositional Reasoning},
  journal = {arXiv preprint arXiv:2603.12266},
  year = {2026},
  url = {https://arxiv.org/abs/2603.12266}
}

@inproceedings{suzgun2023challenging,
  author = {Suzgun, Mirac and Scales, Nathan and Sch{\"a}rli, Nathanael
    and Gehrmann, Sebastian and Tay, Yi and Chung, Hyung Won
    and Chowdhery, Aakanksha and Le, Quoc and Chi, Ed
    and Zhou, Denny and Wei, Jason},
  title = {Challenging {BIG}-Bench Tasks and Whether Chain-of-Thought Can Solve Them},
  booktitle = {Findings of the Association for Computational Linguistics: ACL 2023},
  pages = {13003--13051},
  publisher = {Association for Computational Linguistics},
  year = {2023},
  doi = {10.18653/v1/2023.findings-acl.824},
  url = {https://aclanthology.org/2023.findings-acl.824/}
}

@inproceedings{tafjord2021proofwriter,
  author = {Tafjord, Oyvind and Dalvi, Bhavana and Clark, Peter},
  title = {{ProofWriter}: Generating Implications, Proofs, and Abductive Statements over Natural Language},
  booktitle = {Findings of the Association for Computational Linguistics: ACL-IJCNLP 2021},
  pages = {3621--3634},
  publisher = {Association for Computational Linguistics},
  year = {2021},
  doi = {10.18653/v1/2021.findings-acl.317},
  url = {https://aclanthology.org/2021.findings-acl.317/}
}

@inproceedings{xu2025muslr,
  author = {Xu, Jundong and Fei, Hao and Zhang, Yuhui
    and Pan, Liangming and Huang, Qijun and Liu, Qian
    and Nakov, Preslav and Kan, Min-Yen and Wang, William Yang
    and Lee, Mong-Li and Hsu, Wynne},
  title = {{MuSLR}: Multimodal Symbolic Logical Reasoning},
  booktitle = {Advances in Neural Information Processing Systems},
  year = {2025},
  url = {https://nips.cc/virtual/2025/loc/san-diego/poster/115490}
}

@article{xu2025visulogic,
  author = {Xu, Weiye and Wang, Jiahao and Wang, Weiyun
    and Chen, Zhe and Zhou, Wengang and Yang, Aijun
    and Lu, Lewei and Li, Houqiang and Wang, Xiaohua
    and Zhu, Xizhou and Wang, Wenhai and Dai, Jifeng and Zhu, Jinguo},
  title = {{VisuLogic}: A Benchmark for Evaluating Visual Reasoning in Multi-modal Large Language Models},
  journal = {arXiv preprint arXiv:2504.15279},
  year = {2025},
  url = {https://arxiv.org/abs/2504.15279}
}

@article{zhou2026logicunseen,
  author = {Zhou, Yuchen and Tang, Jiayu and Yang, Shuo
    and Xiao, Xiaoyan and Dai, Yuqin and Yang, Wenhao
    and Gou, Chao and Xia, Xiaobo and Chua, Tat-Seng},
  title = {Logic Unseen: Revealing the Logical Blindspots of Vision-Language Models},
  journal = {Proceedings of the AAAI Conference on Artificial Intelligence},
  volume = {40},
  number = {34},
  year = {2026},
  doi = {10.1609/aaai.v40i34.40143},
  url = {https://ojs.aaai.org/index.php/AAAI/article/view/40143}
}

@book{odonnell2014,
  author = {O'Donnell, Ryan},
  title = {Analysis of Boolean Functions},
  publisher = {Cambridge University Press},
  year = {2014},
  note = {Author's revised version: arXiv:2105.10386},
  url = {https://arxiv.org/abs/2105.10386}
}

@inproceedings{johnson2017clevr,
  author = {Johnson, Justin and Hariharan, Bharath and van der Maaten, Laurens
    and Fei-Fei, Li and Zitnick, C. Lawrence and Girshick, Ross},
  title = {{CLEVR}: A Diagnostic Dataset for Compositional Language and Elementary Visual Reasoning},
  booktitle = {Proceedings of the IEEE Conference on Computer Vision and Pattern Recognition},
  pages = {2901--2910},
  year = {2017},
  url = {https://openaccess.thecvf.com/content_cvpr_2017/html/Johnson_CLEVR_A_Diagnostic_CVPR_2017_paper.html}
}

@inproceedings{hudson2019gqa,
  author = {Hudson, Drew A. and Manning, Christopher D.},
  title = {{GQA}: A New Dataset for Real-World Visual Reasoning and Compositional Question Answering},
  booktitle = {Proceedings of the IEEE/CVF Conference on Computer Vision and Pattern Recognition},
  pages = {6700--6709},
  year = {2019},
  url = {https://arxiv.org/abs/1902.09506}
}

@misc{xiao2024logicvista,
  author = {Xiao, Yijia and Sun, Edward and Liu, Tianyu and Wang, Wei},
  title = {{LogicVista}: Multimodal {LLM} Logical Reasoning Benchmark in Visual Contexts},
  year = {2024},
  note = {arXiv:2407.04973},
  url = {https://arxiv.org/abs/2407.04973}
}

@inproceedings{wei2025satbench,
  title = {{SATBench}: Benchmarking {LLM}s' Logical Reasoning via Automated Puzzle Generation from {SAT} Formulas},
  author = {Wei, Anjiang and Wu, Yuheng and Wan, Yingjia and Suresh, Tarun and Tan, Huanmi and Zhou, Zhanke and Koyejo, Sanmi and Wang, Ke and Aiken, Alex},
  booktitle = {Proceedings of the 2025 Conference on Empirical Methods in Natural Language Processing},
  year = {2025}, pages = {33832--33849},
  publisher = {Association for Computational Linguistics},
  doi = {10.18653/v1/2025.emnlp-main.1716},
  url = {https://aclanthology.org/2025.emnlp-main.1716/}
}

@inproceedings{gokhale2020vqalol,
  author = {Gokhale, Tejas and Banerjee, Pratyay and Baral, Chitta and Yang, Yezhou},
  title = {{VQA-LOL}: Visual Question Answering under the Lens of Logic},
  booktitle = {Computer Vision -- ECCV 2020},
  series = {Lecture Notes in Computer Science},
  volume = {12366},
  pages = {379--396},
  year = {2020},
  publisher = {Springer},
  doi = {10.1007/978-3-030-58589-1_23}
}

@inproceedings{emdin2022parity,
  author = {Emdin, Gregory and Kulikov, Alexander S. and Mihajlin, Ivan and Slezkin, Nikita},
  title = {{CNF} Encodings of Parity},
  booktitle = {47th International Symposium on Mathematical Foundations of Computer Science (MFCS 2022)},
  series = {Leibniz International Proceedings in Informatics},
  volume = {241},
  pages = {47:1--47:12},
  year = {2022},
  publisher = {Schloss Dagstuhl -- Leibniz-Zentrum f{\"u}r Informatik},
  doi = {10.4230/LIPIcs.MFCS.2022.47}
}

@inproceedings{bae2023ehrxqa,
  author = {Bae, Seongsu and Kyung, Daeun and Ryu, Jaehee
    and Cho, Eunbyeol and Lee, Gyubok and Kweon, Sunjun
    and Oh, Jungwoo and Ji, Lei and Chang, Eric I.
    and Kim, Tackeun and Choi, Edward},
  title = {{EHRXQA}: A Multi-Modal Question Answering Dataset
    for Electronic Health Records with Chest X-ray Images},
  booktitle = {Advances in Neural Information Processing Systems},
  volume = {36},
  year = {2023},
  url = {https://proceedings.neurips.cc/paper_files/paper/2023/hash/0c007ebef1d11fd48da6ce4f54687db6-Abstract-Datasets_and_Benchmarks.html}
}

@article{zhang2023equivocal,
  author = {Zhang, Enhao and Daum, Maureen and He, Dong
    and Haynes, Brandon and Krishna, Ranjay
    and Balazinska, Magdalena},
  title = {{EQUI-VOCAL}: Synthesizing Queries for Compositional
    Video Events from Limited User Interactions},
  journal = {Proceedings of the VLDB Endowment},
  volume = {16},
  number = {11},
  pages = {2714--2727},
  year = {2023},
  doi = {10.14778/3611479.3611482},
  url = {https://www.vldb.org/pvldb/vol16/p2714-zhang.pdf}
}

@inproceedings{sima2024drivelm,
  author = {Sima, Chonghao and Renz, Katrin and Chitta, Kashyap
    and Chen, Li and Zhang, Hanxue and Xie, Chengen
    and Bei{\ss}wenger, Jens and Luo, Ping
    and Geiger, Andreas and Li, Hongyang},
  title = {{DriveLM}: Driving with Graph Visual Question Answering},
  booktitle = {European Conference on Computer Vision},
  year = {2024},
  url = {https://arxiv.org/abs/2312.14150}
}

@inproceedings{merrill2024expressive,
  author = {Merrill, William and Sabharwal, Ashish},
  title = {The Expressive Power of Transformers with Chain of Thought},
  booktitle = {International Conference on Learning Representations},
  year = {2024},
  url = {https://proceedings.iclr.cc/paper_files/paper/2024/hash/1f59721c106ea80f613299039112f651-Abstract-Conference.html}
}

@inproceedings{li2024serial,
  author = {Li, Zhiyuan and Liu, Hong and Zhou, Denny and Ma, Tengyu},
  title = {Chain of Thought Empowers Transformers
    to Solve Inherently Serial Problems},
  booktitle = {International Conference on Learning Representations},
  year = {2024},
  url = {https://proceedings.iclr.cc/paper_files/paper/2024/hash/3309b4112c9f04a993f2bbdd0274bba1-Abstract-Conference.html}
}

@inproceedings{shah2019cycle,
  author = {Shah, Meet and Chen, Xinlei and Rohrbach, Marcus
    and Parikh, Devi},
  title = {Cycle-Consistency for Robust Visual Question Answering},
  booktitle = {Proceedings of the IEEE/CVF Conference
    on Computer Vision and Pattern Recognition},
  year = {2019},
  url = {https://arxiv.org/abs/1902.05660}
}

@misc{zhang2024goodcaptioningbadcounting,
      title={Good at captioning, bad at counting: Benchmarking GPT-4V on Earth observation data}, 
      author={Chenhui Zhang and Sherrie Wang},
      year={2024},
      eprint={2401.17600},
      archivePrefix={arXiv},
      primaryClass={cs.CL},
      url={https://arxiv.org/abs/2401.17600}, 
}

@book{demorgan1847formallogic,
  title     = {Formal Logic, or, The Calculus of Inference, Necessary and Probable},
  author    = {De Morgan, Augustus},
  year      = {1847},
  publisher = {Taylor and Walton},
  address   = {London, England}
}

@Article{su14137781,
AUTHOR = {Liu, Huwei and Zhou, Li and Zhao, Junhui and Wang, Fan and Yang, Jianglong and Liang, Kaibo and Li, Zhaochan},
TITLE = {Deep-Learning-Based Accurate Identification of Warehouse Goods for Robot Picking Operations},
JOURNAL = {Sustainability},
VOLUME = {14},
YEAR = {2022},
NUMBER = {13},
ARTICLE-NUMBER = {7781},
URL = {https://www.mdpi.com/2071-1050/14/13/7781},
ISSN = {2071-1050},
DOI = {10.3390/su14137781}
}

@INPROCEEDINGS{10448946,
  author={Subramanian, M and Agalya, M and Divya, V and Harini, R},
  booktitle={2023 Intelligent Computing and Control for Engineering and Business Systems (ICCEBS)}, 
  title={AI Based Warehouse Management System}, 
  year={2023},
  volume={},
  number={},
  pages={1-5},
  doi={10.1109/ICCEBS58601.2023.10448946}}

@misc{soler2024combiningdeeplearningstreet,
      title={Combining Deep Learning and Street View Imagery to Map Smallholder Crop Types}, 
      author={Jordi Laguarta Soler and Thomas Friedel and Sherrie Wang},
      year={2024},
      eprint={2309.05930},
      archivePrefix={arXiv},
      primaryClass={cs.CV},
      url={https://arxiv.org/abs/2309.05930}, 
}

@Article{rs14225738,
AUTHOR = {Wang, Sherrie and Waldner, François and Lobell, David B.},
TITLE = {Unlocking Large-Scale Crop Field Delineation in Smallholder Farming Systems with Transfer Learning and Weak Supervision},
JOURNAL = {Remote Sensing},
VOLUME = {14},
YEAR = {2022},
NUMBER = {22},
ARTICLE-NUMBER = {5738},
URL = {https://www.mdpi.com/2072-4292/14/22/5738},
ISSN = {2072-4292},
DOI = {10.3390/rs14225738}
}

@misc{anand2026smatattentionstructuredlongcontextsequence,
      title={{SMat-Attention: Structured Long-Context Sequence Modeling}}, 
      author={Emile Anand and Abdullah Ateyeh and Archer Wang and Marin Soljačić},
      year={2026},
      eprint={2609.36062},
      archivePrefix={arXiv},
      primaryClass={cs.AI},
      url={https://arxiv.org/abs/2609.36062}, 
}

@misc{chang2026globalpiqaevaluatingcommonsense,
      title={Global PIQA: Evaluating Commonsense Reasoning Across 100+ Languages and Cultures}, 
      author={Tyler A. Chang and Catherine Arnett and Authors at the 5th Multilingual Representation Learning (MRL) Workshop},
      year={2026},
      eprint={2510.24081},
      archivePrefix={arXiv},
      primaryClass={cs.CL},
      url={https://arxiv.org/abs/2510.24081}, 
}

@inproceedings{lee-etal-2026-fol,
    title = "{FOL}-Traces: Verified First-Order Logic Reasoning Traces at Scale",
    author = "Lee, Isabelle  and
      Liaw, Sarah  and
      Yogatama, Dani",
    editor = "Demberg, Vera  and
      Inui, Kentaro  and
      Marquez, Llu{\'i}s",
    booktitle = "Findings of the {A}ssociation for {C}omputational {L}inguistics: {EACL} 2026",
    month = mar,
    year = "2026",
    address = "Rabat, Morocco",
    publisher = "Association for Computational Linguistics",
    url = "https://aclanthology.org/2026.findings-eacl.115/",
    doi = "10.18653/v1/2026.findings-eacl.115",
    pages = "2181--2203",
    ISBN = "979-8-89176-386-9"
}

@misc{anand2026continuouslatentcontextsenable,
      title={Continuous Latent Contexts Enable Efficient Online Learning in Transformers}, 
      author={Emile Anand and Abdullah Ateyeh and Xinyuan Cao and Max Dabagia},
      year={2026},
      eprint={2605.09867},
      archivePrefix={arXiv},
      primaryClass={cs.LG},
      url={https://arxiv.org/abs/2605.09867}, 
}
\clearpage
\appendix
\section{Inference Settings and Additional Results}
\label{app:experiments}
\label{app:experiment-details}

\subsection{Inference Configurations, Prompting, and Scoring}

\paragraph{Recorded configurations.}
Table~\ref{tab:recorded-api-configurations} lists the requested API model identifiers, providers, thinking settings, and output-token caps. For each configuration, these settings are the same across the four task--format conditions. Identifiers are recorded as requested and do not
independently establish immutable backend versions; in particular, the DeepSeek requests use the alias \texttt{deepseek-flash}.

\begin{table}[htbp]
\centering\small
\setlength{\tabcolsep}{3pt}
\renewcommand{\arraystretch}{1.1}
\begin{tabularx}{\linewidth}{@{}l>{\raggedright\arraybackslash}Xlrl@{}}
\toprule
Provider & Requested model identifier & Mode & Cap & Init. dates \\
\midrule
OpenAI & \path{gpt-5.6-luna} & none & 2,048 & 14 \\
Anthropic & \path{claude-haiku-4-5-20251001} & disabled & 2,048 & 14 \\
xAI & \path{grok-4.3} & none & 2,048 & 14 \\
Google & \path{gemini-3.5-flash-lite} & minimal & 2,048 & 14--15 \\
DeepSeek & \path{deepseek-flash} & disabled & 2,048 & 15 \\
Z.ai & \path{glm-4.6v-flashx} & disabled & 2,048 & 15 \\
DeepInfra & \path{meta-llama/Llama-4-Scout-17B-16E-Instruct} & Instruct & 2,048 & 15 \\
Alibaba & \path{qwen3-vl-8b-instruct} & Instruct & 2,048 & 15--16 \\
Z.ai & \path{glm-4.6v-flashx} & enabled & 4,096 & 15--19 \\
\bottomrule
\end{tabularx}
\caption{Recorded inference configurations. Mode denotes the requested thinking setting or selection of an Instruct variant; it is not a measurement of internal reasoning. Cap denotes the requested output-token limit, including
thinking tokens where the provider counts them toward that limit. Initialization dates are days in September 2026 (UTC), spanning creation of the four task--format run manifests; they are not completion or retry dates.}
\label{tab:recorded-api-configurations}
\end{table}

\paragraph{Provider-specific request fields.}
For OpenAI, \texttt{none} is sent as \path{reasoning.effort=none}; for xAI,
it is sent as \path{reasoning_effort=none}. Anthropic, DeepSeek, and thinking-disabled GLM receive \path{thinking.type=disabled}; thinking-enabled GLM receives \path{thinking.type=enabled}. Gemini receives
\path{generationConfig.thinkingConfig.thinkingLevel=minimal}. For Qwen and Llama, the harness records \texttt{none} to identify the selected
Instruct variant, but sends no thinking-toggle parameter. OpenAI, xAI, and DeepSeek receive image detail \texttt{high}; that parameter is omitted for the other configurations. Anthropic requests use the \texttt{anthropic-version} header \texttt{2023-06-01}.

\paragraph{Provider endpoints.}
The requests use the following API roots and routes: 
\begin{itemize}[leftmargin=*,nosep]
    \item OpenAI: \url{https://api.openai.com/v1}, 
    route \path{/responses}. 
    \item Anthropic: \url{https://api.anthropic.com/v1}, 
    route \path{/messages}. 
    \item xAI: 
    \url{https://api.x.ai/v1}, 
    route \path{/chat/completions}.
    \item Google: \url{https://generativelanguage.googleapis.com/v1beta}, 
    route \path{/models/gemini-3.5-flash-lite:generateContent}.
    \item DeepSeek: \url{https://api.deepseek.com}, 
    route \path{/chat/completions}.
    \item Z.ai: \url{https://api.z.ai/api/paas/v4}, 
    route \path{/chat/completions}.
    \item DeepInfra: \url{https://api.deepinfra.com/v1/openai}, route \path{/chat/completions}.
    \item Alibaba Model Studio: workspace-specific root \path{https://WORKSPACE.ap-southeast-1.maas.aliyuncs.com/compatible-mode/v1},
    route \path{/chat/completions}. \texttt{WORKSPACE} replaces the account-specific host prefix for anonymous reporting; the region and API path are retained.
\end{itemize}

\paragraph{Sampling and dates.}
Temperature, top-$p$, top-$k$, and a sampling seed are not explicitly set in the requests. The manifests record temperature as null, which the request builder interprets as omission rather than a numeric setting. The provider-default numerical values were not retained, so these runs should not be described as temperature-zero or deterministic evaluations.
The manifests record UTC creation times between September 14 and 19, 2026. Per-request wall-clock timestamps and exact completion times of later retries were not retained; the initialization dates in Table~\ref{tab:recorded-api-configurations} should not be interpreted as complete evaluation intervals.

\textbf{Prompt policies.}
GPT-5.6 Luna uses the original dataset prompts in all four task--format conditions. Every other Boolean configuration appends: ``Respond with exactly one word: Yes or No. Do not include explanations, Markdown, punctuation, or any other text.'' For identification, Claude Haiku 4.5, DeepSeek, both GLM configurations, Llama, and Qwen append a label-only instruction; Grok and Gemini use the original prompts. The recorded policies are \texttt{original}, \texttt{boolean-one-word-v1}, and \texttt{answer-only-v1}. The original identification prompt already requests a single label, which the suffix reinforces. Prepared-image hashes verify that matched questions use the same image. Prompt hashes identify the format-specific prompts and verify that the same prompt is used in both GLM thinking configurations.

\textbf{Parsers and retries.}
Case, surrounding whitespace, and one terminal period or exclamation mark are tolerated. Luna's saved policy accepts a plain answer; all other configurations additionally accept one answer enclosed in Markdown bold. Uniformly reapplying the plain-or-bold parser to completed outputs changes none of the reported accuracies. Multiple labels, explanatory prose, empty final answers, and incomplete generations remain incorrect, even if they mention the reference answer. Only infrastructure errors are retried; the most recent settled response is scored once per ID. Earlier error rows remain in the append-only histories. 

\subsection{Equivalent-Expression Outcomes}

For each of the $M=3{,}000$ base--De Morgan pairs, let $\hat{y}_{j,0}$ and $\hat{y}_{j,1}$ be the model's final predictions and let $y_j$ be their shared reference label. Let $V$ be the set of pairs with two valid predictions. We define
\[
\mathrm{BC}
=\frac{1}{M}\sum_{j=1}^{M}
\mathbf{1}\!\left[\hat{y}_{j,0}=\hat{y}_{j,1}=y_j\right],
\qquad
\mathrm{Conflict}
=\frac{1}{|V|}\sum_{j\in V}
\mathbf{1}\!\left[\hat{y}_{j,0}\ne\hat{y}_{j,1}\right].
\]
Invalid predictions never count as correct. Both-correct accuracy uses all $M$ pairs, whereas conflict uses only the $n_v=|V|$ pairs with two valid outputs. We report  these valid-pair denominators. Because the reference labels are balanced, a constant-label predictor has $50\%$ both-correct accuracy and zero conflict. Both metrics must therefore be interpreted alongside item accuracy and predicted-label frequencies.

\begin{table}[htbp]
\centering\small
\setlength{\tabcolsep}{4pt}

\begin{tabular}{@{}lrrrrrr@{}}
\toprule
Configuration & Sym. BC & Conflict & $n_v$ & NL BC & Conflict & $n_v$ \\
\midrule
Luna & 35.27 & 27.55 & 2998 & 41.83 & 13.67 & 3000 \\
Haiku 4.5 & 47.70 & 2.44 & 2945 & 44.67 & 10.92 & 2985 \\
Grok 4.3 & 38.67 & 20.17 & 3000 & 31.37 & 36.40 & 3000 \\
Gemini Flash-Lite & 37.57 & 22.07 & 3000 & 35.83 & 25.37 & 3000 \\
DeepSeek Flash & 47.63 & 3.27 & 2936 & 47.93 & 3.44 & 2997 \\
GLM FlashX (off) & 39.13 & 18.90 & 3000 & 49.53 & 1.67 & 3000 \\
Llama 4 Scout & 46.83 & 6.30 & 3000 & 41.50 & 14.87 & 3000 \\
Qwen3-VL 8B & 41.10 & 18.93 & 3000 & 41.87 & 17.00 & 3000 \\
\midrule
GLM FlashX (on) & 45.13 & 36.47 & 2953 & 32.60 & 34.37 & 2502 \\
\bottomrule
\end{tabular}
\caption{Equivalent-expression outcomes. BC (both-correct accuracy, \%) uses all 3,000 base--De Morgan pairs per format; conflict (\%) uses only pairs with two valid outputs, counted by $n_v$. Invalid outputs never count as correct. A constant-label predictor achieves $50\%$ BC and zero conflict; interpret these metrics alongside item accuracy and label frequencies.}
\end{table}

For thinking-enabled GLM, natural-language both-correct accuracy is $32.60\%$, and conflict is $34.37\%$ among the $2{,}502$ pairs with two valid outputs. This reduced valid-pair coverage accompanies many token-limit failures in the natural-language condition. Several other configurations combine low conflict with strongly skewed label frequencies (Table~\ref{tab:label-bias}). For example, DeepSeek returns a valid Yes answer on only $1.78\%$ of all symbolic Boolean questions, with $3.27\%$ conflict and $49.62\%$ item accuracy.

\subsection{Validity, Stratified Accuracy, and Cross-Format Outcomes}

Table~\ref{tab:invalid-outputs} reports invalid responses and identifies the subset that terminates at the token limit. Thinking-enabled GLM produces $827$ invalid responses: $826$ token-limit failures and one incomplete response with an empty final answer. Of its $47$ invalid symbolic Boolean responses, $46$ reach the token limit; all $536$ invalid natural-language Boolean responses do so. DeepSeek produces $639$ invalid responses, including $81$ token-limit failures, and Gemini produces $184$, including $10$ token-limit failures. Their remaining invalid responses finish generation but fail the final-answer parser. Every invalid response remains in the accuracy denominator.

\begin{table}[H]
\centering\small
\setlength{\tabcolsep}{4pt}

\begin{tabular}{@{}lrrrrr@{}}
\toprule
Configuration & Bool. sym. & Bool. NL & Ident. sym. & Ident. NL & Token limit \\
\midrule
GPT-5.6 Luna & 2 & 0 & 0 & 0 & 0 \\
Claude Haiku 4.5 & 57 & 15 & 31 & 44 & 0 \\
Grok 4.3 & 0 & 0 & 13 & 12 & 0 \\
Gemini 3.5 Flash-Lite & 0 & 0 & 183 & 1 & 10 \\
DeepSeek Flash & 72 & 3 & 317 & 247 & 81 \\
GLM-4.6V-FlashX (off) & 0 & 0 & 0 & 0 & 0 \\
Llama 4 Scout & 0 & 0 & 0 & 0 & 0 \\
Qwen3-VL-8B-Instruct & 0 & 0 & 1 & 0 & 1 \\
\midrule
GLM-4.6V-FlashX (on) & 47 & 536 & 144 & 100 & 826 \\
\bottomrule
\end{tabular}
\caption{Invalid-response counts for each task--format condition. Each Boolean column contains $6{,}000$ questions, and each identification column contains $1{,}000$. The final column counts token-limit failures across all four conditions; these are included in the preceding invalid-response counts. Every question has a final response after retries. Completed responses containing extra prose and responses cut off at the token limit are both scored as invalid, but represent different failure modes.}
\label{tab:invalid-outputs}
\end{table}
\begin{table}[H]
\centering\small
\setlength{\tabcolsep}{4pt}

\begin{tabular}{@{}lrrrrr@{}}
\toprule
Family & $N$ & Sym. off & Sym. on & NL off & NL on \\
\midrule
XOR of conjunctions & 1000 & 50.60 & 56.80 & 51.80 & 47.30 \\
XOR of majorities & 1000 & 47.50 & 52.00 & 47.80 & 44.50 \\
Nested multiplexers & 1500 & 50.60 & 80.53 & 52.33 & 58.53 \\
Dense CNF & 1000 & 45.90 & 58.40 & 47.90 & 47.80 \\
Sparse CNF control & 500 & 40.80 & 86.20 & 52.00 & 67.60 \\
Polarity trap + parity & 500 & 53.40 & 47.40 & 53.60 & 49.60 \\
Parity control & 500 & 49.00 & 52.40 & 46.80 & 42.40 \\
\bottomrule
\end{tabular}
\caption{GLM-4.6V-FlashX Boolean accuracy (\%) by construction family. $N$ is the number of questions per format. Off and on denote the thinking-disabled and thinking-enabled configurations, with output-token caps of $2{,}048$ and $4{,}096$, respectively. }
\label{tab:glm-cohorts}
\end{table}
\begin{table}[H]
\centering\small
\setlength{\tabcolsep}{4pt}

\begin{tabular}{@{}lrrrr@{}}
\toprule
Configuration & Sym. synth. & Sym. photo & NL synth. & NL photo \\
\midrule
GPT-5.6 Luna & 49.15 & 48.85 & 48.23 & 49.55 \\
Claude Haiku 4.5 & 49.03 & 50.05 & 50.10 & 50.45 \\
Grok 4.3 & 48.48 & 49.30 & 49.40 & 49.90 \\
Gemini 3.5 Flash-Lite & 48.08 & 49.65 & 48.15 & 49.25 \\
DeepSeek Flash & 50.00 & 48.85 & 49.55 & 49.90 \\
GLM-4.6V-FlashX (off) & 48.13 & 49.50 & 50.10 & 50.90 \\
Llama 4 Scout & 49.88 & 50.20 & 48.23 & 50.35 \\
Qwen3-VL-8B-Instruct & 50.10 & 51.50 & 49.45 & 52.20 \\
\midrule
GLM-4.6V-FlashX (on) & 65.18 & 60.15 & 51.68 & 50.25 \\
\bottomrule
\end{tabular}
\caption{Boolean accuracy (\%) by image source, using $4{,}000$ questions over generated scenes and $2{,}000$ questions over photographs per format. Photographic scores are computed against the supplied reference annotations.}
\label{tab:track-results}
\end{table}
\begin{table}[H]
\centering\small
\setlength{\tabcolsep}{4pt}

\begin{tabular}{@{}lrr@{}}
\toprule
Configuration & Symbolic Yes (\%) & NL Yes (\%) \\
\midrule
GPT-5.6 Luna & 74.98 & 89.10 \\
Claude Haiku 4.5 & 1.35 & 13.62 \\
Grok 4.3 & 13.12 & 29.87 \\
Gemini 3.5 Flash-Lite & 81.23 & 80.08 \\
DeepSeek Flash & 1.78 & 2.08 \\
GLM-4.6V-FlashX (off) & 78.32 & 98.60 \\
Llama 4 Scout & 96.25 & 51.20 \\
Qwen3-VL-8B-Instruct & 26.00 & 63.27 \\
\midrule
GLM-4.6V-FlashX (on) & 54.13 & 62.58 \\
\bottomrule
\end{tabular}
\caption{Valid Yes responses as a percentage of all $6{,}000$ Boolean questions in each format. Reference labels are exactly balanced. The remaining percentage includes both No responses and invalid outputs; Table~\ref{tab:invalid-outputs} reports invalid-output counts.}
\label{tab:label-bias}
\end{table}
\begin{table}[H]
\centering\small
\setlength{\tabcolsep}{4pt}

\begin{tabular}{@{}lrrrrrr@{}}
\toprule
Configuration & Bool. BC & S-only & N-only & Ident. BC & S-only & N-only \\
\midrule
GPT-5.6 Luna & 35.52 & 812 & 789 & 27.20 & 85 & 128 \\
Claude Haiku 4.5 & 42.73 & 398 & 449 & 27.20 & 50 & 158 \\
Grok 4.3 & 33.12 & 938 & 987 & 13.50 & 81 & 122 \\
Gemini 3.5 Flash-Lite & 35.75 & 771 & 766 & 21.50 & 56 & 196 \\
DeepSeek Flash & 47.53 & 125 & 128 & 8.30 & 82 & 160 \\
GLM-4.6V-FlashX (off) & 38.12 & 628 & 735 & 24.10 & 72 & 109 \\
Llama 4 Scout & 26.67 & 1399 & 1336 & 22.80 & 50 & 105 \\
Qwen3-VL-8B-Instruct & 24.87 & 1542 & 1530 & 19.40 & 36 & 92 \\
\midrule
GLM-4.6V-FlashX (on) & 35.62 & 1673 & 935 & 63.20 & 105 & 89 \\
\bottomrule
\end{tabular}
\caption{Matched symbolic and natural-language outcomes for the same question IDs. Here, BC denotes the percentage of questions answered correctly in both formats, rather than both members of an equivalent-expression pair. S-only and N-only count questions answered correctly only in the symbolic or natural-language format, respectively. The Boolean comparison contains $6{,}000$ matched questions, and the identification comparison contains $1{,}000$. Invalid outputs count as incorrect.}
\label{tab:format-pairs}
\end{table}

\subsection{Performance by logical properties}
\label{app:logical-property-results}

\paragraph{Scope and metrics.}
We join the saved Boolean responses to the logical annotations by question ID,
covering $6{,}000$ questions in each format for all nine configurations.
The analysis uses existing responses without additional inference.
Certificate size $c_F(a)$ is defined in Equation~\ref{eq:certificate};
$A_2(F)$ is the optimal fixed-two-input predictor's accuracy over the uniform
Boolean domain, as defined in Appendix~\ref{app:construction}.
We summarize input sensitivity by total influence,
$I(F)=\sum_{i=1}^{10}\operatorname{Inf}_i(F)$.
Neither $A_2(F)$ nor $I(F)$ measures a model's visual-fact accuracy or identifies
its internal decision procedure.

\begin{figure}[htbp]
\centering
\includegraphics[width=\linewidth]{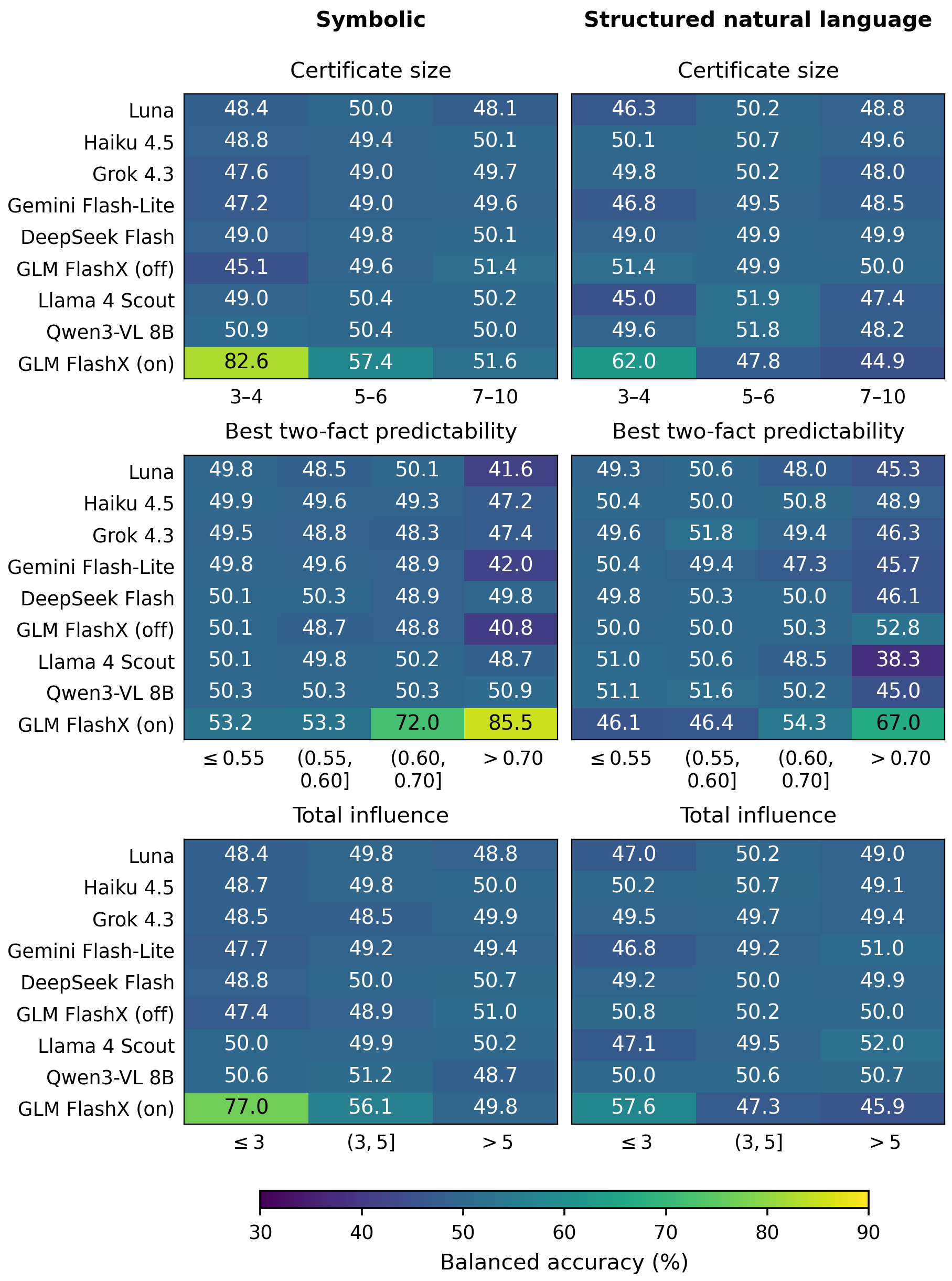}
\caption{Balanced Boolean accuracy (\%) by logical-property group. Columns show symbolic and structured natural-language formats; rows show certificate size, best two-fact predictability, and total influence. Groups differ in their construction-family
composition, so the figure reports descriptive associations rather than isolated effects of each property.}
\label{fig:property-accuracy}
\end{figure}

Figure~\ref{fig:property-accuracy} reports balanced accuracy, the mean of
Yes-label and No-label recall within each subgroup. Invalid responses count
as incorrect. This metric gives a constant-label predictor a $50\%$ baseline
even when a subgroup's reference labels are imbalanced. The main-text table
uses ordinary accuracy to match the other reported results. For thinking-enabled
GLM's symbolic certificate groups, the balanced accuracies are $82.62\%$,
$57.44\%$, and $51.60\%$, close to the ordinary accuracies in
Table~\ref{tab:property-certificate}.

The certificate groups contain $1{,}772$, $2{,}856$, and $1{,}372$ questions;
the four $A_2$ groups contain $2{,}000$, $1{,}030$, $2{,}488$, and $482$;
the three total-influence groups contain $2{,}436$, $2{,}564$, and $1{,}000$.
These counts are per format and include both equivalent-expression variants.
The exploratory bins were chosen before inspecting model scores, and exact
certificate-size results are also reported below. Questions share images and
logical functions; we make no independent-trial significance claims.

\paragraph{Predictability, influence, and family overlap.}
Thinking-enabled GLM's ordinary symbolic accuracy is $53.35\%$ for
$A_2\leq0.55$ and $85.48\%$ for $A_2>0.70$; it is $76.97\%$ for $I(F)\leq3$
and $49.90\%$ for $I(F)>5$. The other eight configurations do not show a
comparable, consistent advantage on these groups (Figure~\ref{fig:property-accuracy}).
However, Table~\ref{tab:property-support} shows that $A_2$ and total influence
vary within only the dense and sparse CNF families. Most variation in these
properties therefore corresponds to differences between construction families.
The observed associations do not establish that a model relies on two facts
or that sensitivity to visual-fact errors explains its failures.

\begin{table}[htbp]
\centering\small
\begin{tabular}{@{}lrrrr@{}}
\toprule
Family & $N$ & $c_F(a)$ & $A_2(F)$ & $I(F)$ \\
\midrule
Parity of conjunctions & 1,000 & 5--10 & 0.5312 & 5.0000 \\
Parity of majorities & 1,000 & 6 & 0.5703 & 3.7500 \\
Nested multiplexers & 1,500 & 4--5 & 0.6250 & 2.8594 \\
Dense CNF & 1,000 & 4--8 & 0.5889--0.7158 & 2.5508--3.3047 \\
Sparse chain CNF & 500 & 3--6 & 0.6816--0.8350 & 1.7656--2.4766 \\
Polarity-trap composition & 500 & 7--10 & 0.5000 & 6.5000 \\
Ten-input parity & 500 & 10 & 0.5000 & 10.0000 \\
\bottomrule
\end{tabular}
\caption{Observed logical-property ranges within construction families.
$N$ counts questions per format. A single value indicates no within-family
variation; influence and predictability are rounded to four decimal places.
Ranges describe the released questions, not all possible formulas in a family.}
\label{tab:property-support}
\end{table}

\paragraph{Within-family comparisons.}
Table~\ref{tab:certificate-within-family} reports thinking-enabled GLM's
ordinary accuracy at every observed certificate size in the five families
where certificate size varies. Nested-multiplexer symbolic accuracy is
$81.25\%$ at size four and $75.00\%$ at size five. Sparse-CNF symbolic accuracy
decreases from sizes three through five, but rises at size six, a group
containing only 22 questions (11 equivalent-expression pairs).
Parity of conjunctions also exhibits reversals. These comparisons support
using the annotations to characterize performance, while cautioning against
a universal monotonic relationship between certificate size and difficulty.

\begin{table}[htbp]
\centering\small
\begin{tabular}{@{}rrrr@{}}
\toprule
Certificate & $N$ & Symbolic & NL \\
\midrule
\multicolumn{4}{l}{\emph{Parity of conjunctions}} \\
5 & 232 & 74.14 & 51.72 \\
6 & 436 & 47.71 & 48.17 \\
7 & 238 & 62.18 & 39.92 \\
8 & 74 & 43.24 & 56.76 \\
9 & 14 & 50.00 & 21.43 \\
10 & 6 & 16.67 & 50.00 \\
\addlinespace
\multicolumn{4}{l}{\emph{Nested multiplexers}} \\
4 & 1,328 & 81.25 & 58.96 \\
5 & 172 & 75.00 & 55.23 \\
\addlinespace
\multicolumn{4}{l}{\emph{Dense CNF}} \\
4 & 38 & 65.79 & 52.63 \\
5 & 718 & 57.80 & 49.30 \\
6 & 204 & 60.29 & 44.61 \\
7 & 38 & 52.63 & 34.21 \\
8 & 2 & 50.00 & 0.00 \\
\addlinespace
\multicolumn{4}{l}{\emph{Sparse chain CNF}} \\
3 & 330 & 90.61 & 77.27 \\
4 & 76 & 80.26 & 57.89 \\
5 & 72 & 69.44 & 43.06 \\
6 & 22 & 95.45 & 36.36 \\
\addlinespace
\multicolumn{4}{l}{\emph{Polarity-trap composition}} \\
7 & 486 & 47.74 & 50.41 \\
10 & 14 & 35.71 & 21.43 \\
\addlinespace
\bottomrule
\end{tabular}
\caption{Thinking-enabled GLM accuracy (\%) by certificate size within each
family that exhibits certificate variation. $N$ counts questions per format;
each equivalent-expression pair contributes two questions. Invalid outputs
remain incorrect. Small groups, including those with fewer than 50 questions,
should be interpreted cautiously.}
\label{tab:certificate-within-family}
\end{table}

\textbf{Offline regeneration.}
Run \texttt{python -m evals.paper\_results} from the repository root. The generator reads the run manifests, prediction histories, and final summaries; checks question-ID coverage and retry histories; recomputes the metrics; and verifies the required matches in reference labels, prompts, and prepared images. It writes \texttt{paper/generated/results.json}, which records exact numerators and denominators, inference settings, invalid-response finish reasons, and source SHA-256 hashes. The reported evaluation contains $36$ complete task--format runs and $126{,}000$ final responses. Regenerating the tables requires no model API calls and leaves the raw histories unchanged.

\section{Construction Details and Diagnostic Scope}
\label{app:construction}
\label{sec:generation}
\label{app:generation}
\label{sec:grounding}
\label{sec:photographs}

\subsection{Expression sampling and rewrite accounting}

The seven construction families in Table~\ref{tab:families} are instantiated with shuffled variable order and random literal signs. Accepted bases are nonconstant, have ten essential inputs, and satisfy $q(F)\in[0.1,0.9]$. Negating an entire expression changes its answer without changing its input influences or certificate sizes. Family membership, generation seeds, both expression trees, rewrite traces, and reference assignments are recorded with each question.

The released rewrite rules are
\begin{equation}
 A\land B\equiv\neg(\neg A\lor\neg B),\qquad
 A\lor B\equiv\neg(\neg A\land\neg B).
 \label{eq:demorgan}
\end{equation}

Each rewrite swaps the selected AND/OR operator and introduces three NOT nodes; the child subexpressions are retained. At most three disjoint rewrites are applied. If the base AST has $N_0$ nodes, the total number of added nodes cannot exceed $\lfloor0.15N_0\rfloor$. We check all 1,024 assignments to verify that each rewritten expression has the same truth table as its base. The resulting 3,000 pairs also share their image, visual statements, and reference answer. Double negation, duplication, commutative reordering, and visual-layout changes are not separate experimental conditions in this release.

\subsection{Construction of the seven expression families}
\label{app:family-construction}

The families in Table~\ref{tab:families} specify how ten image-grounded statements are combined. This subsection gives their construction templates, explains their evaluation, and distinguishes their intended diagnostic roles. The worked assignments below illustrate the templates, where we write $H$ for a template before the optional whole-expression negation used for answer balancing, so the released base expression is either $F=H$ or $F=\neg H$.

\textbf{Notation and shared construction steps.}
A literal is a variable or its negation. Let $\pi$ be a randomly sampled permutation of the ten variable indices. For the parity, majority, multiplexer, and polarity-trap templates, define $\ell_i$ to be either $x_{\pi(i)}$ or $\neg x_{\pi(i)}$, with each sign chosen independently with probability $1/2$. Repeated occurrences of $\ell_i$ use the same signed variable. In the two CNF families, signs are instead sampled independently for each occurrence within a clause; the same variable may therefore appear with different signs in different clauses. Unless stated otherwise, the example values refer to the signed literals, after applying their negations.

The symbol $\oplus$ denotes exclusive OR for two inputs and odd parity for multiple inputs. A multiplexer is a conditional choice,
\[
 M(s,u,v)=(s\land u)\lor(\neg s\land v):
 \quad M(1,u,v)=u,\qquad M(0,u,v)=v.
\]
For an odd number $m$ of inputs, $\mathrm{Maj}_m$ is true when at least $(m+1)/2$ inputs are true. These are construction abbreviations. The generator expands each XOR as $(u\land\neg v)\lor(\neg u\land v)$, each multiplexer as the expression above, and each majority as an OR of all conjunctions of $(m+1)/2$ distinct inputs. Multi-input AND, OR, and XOR operations are combined by recursively splitting their ordered input lists into two nearly equal parts. The displayed expressions consequently contain only binary AND/OR and unary NOT; neither named modules nor intermediate variables are supplied to models.

All families share rejection filters: the true-output fraction over the $1{,}024$ assignments must lie in $[0.1,0.9]$, and every variable must have influence at least $0.02$. Family-specific checks are described below. After acceptance, visual statements are randomly assigned to the variable names, and the entire expression may be negated to satisfy the image's assigned Yes/No quota. Each accepted base receives a verified De Morgan partner. The fixed family schedule yields $500$, $500$, $750$, $500$, $250$, $250$, and $250$ bases in the order below; doubling these counts gives the $6{,}000$ questions in Table~\ref{tab:families}. 

\subsubsection{Parity of conjunctions}
\label{app:family-conjunctions}

\textbf{Meaning and construction.}
The generator partitions the ten signed literals into five disjoint pairs, evaluates whether both members of each pair are true, and takes the parity of those five results:
\[
 t_j=\ell_{2j-1}\land\ell_{2j}\quad(j=1,\ldots,5),
 \qquad H=t_1\oplus t_2\oplus t_3\oplus t_4\oplus t_5.
\]
The answer before output balancing is true when exactly one, three, or five pairs are jointly true. Variable permutation changes which visual statements are paired, and literal signs determine whether each pair requires a statement to hold or fail. In the implementation, this family is named \texttt{quadratic\_parity}; the name reflects its five degree-two terms over Boolean arithmetic.

\textbf{Worked evaluation and diagnostic role.}
Consider signed values $(1,1,1,1,1,0,0,0,0,0)$. The pair results are $(1,1,0,0,0)$, so $H=0$, even though five individual literals are true. Changing the sixth value to one makes the pair results $(1,1,1,0,0)$ and changes $H$ to one. Thus counting true visual statements alone does not implement this function: the model must preserve the pairing and combine intermediate conjunctions. This family supports analysis of dependence on multiple facts and prediction from small subsets; the corresponding influence, certificate, and two-statement-prediction measurements are computed on each released function rather than inferred from its name.

\subsubsection{Parity of majorities}
\label{app:family-majorities}

\textbf{Meaning and construction.}
The ten signed literals are divided into two disjoint groups of five. Each group produces a majority decision, and the complete condition is true exactly when those decisions differ:
\[
 H=\mathrm{Maj}_5(\ell_1,\ldots,\ell_5)
 \oplus\mathrm{Maj}_5(\ell_6,\ldots,\ell_{10}).
\]
Each majority requires at least three true literals. Its expansion contains the ten possible three-literal conjunctions, joined by OR; four or five true literals satisfy multiple such conjunctions without changing the majority result. Random permutation determines the two groups, while the signs determine which positive or negative visual facts count toward each threshold.

\textbf{Worked evaluation and diagnostic role.}
Suppose the two groups contain three and one true literals. Their majority values are $(1,0)$, giving $H=1$. If they instead contain two true literals each, the majority values are $(0,0)$ and $H=0$, despite the same total of four true literals. The grouping therefore matters, as do the two threshold decisions. This family examines composition of intermediate decisions and contrasts with both pairwise conjunctions and direct parity. Therefore, its symbolic length also reflects expansion of majority into the permitted connectives.

\subsubsection{Nested multiplexers with shared inputs}
\label{app:family-multiplexers}

\textbf{Meaning and construction.}
A multiplexer uses its first input to choose which of two remaining conditions determines its value. The generator nests these choices and reuses some signed literals in different computational roles. Its exact template is
\[
\begin{aligned}
 A&=\ell_4\oplus\ell_5,
 &B&=\mathrm{Maj}_3(\ell_6,\ell_7,\ell_8),\\
 C&=\ell_9\oplus\ell_{10},
 &D&=M\bigl(\ell_5,\ell_4\oplus\ell_9,
                  \ell_7\mathbin{\circ}\ell_{10}\bigr),\\
 H&=M\bigl(\ell_1,M(\ell_2,A,B),M(\ell_3,C,D)\bigr),
\end{aligned}
\]
where $\circ$ is sampled as AND or OR with equal probability. The symbols $A,B,C,D$ name subexpressions only for exposition. For instance, $\ell_5$ is both an input to $A$ and the selector of $D$, while $\ell_9$ contributes to both $C$ and one branch of $D$. A repeated statement retains its truth value wherever it appears.

\textbf{Worked evaluation and diagnostic role.}
If $\ell_1=0$, the outer choice selects $M(\ell_3,C,D)$. If also $\ell_3=0$ and $\ell_5=1$, evaluation selects $D$ and then $\ell_4\oplus\ell_9$. Setting $\ell_4=0$ and $\ell_9=1$ therefore fixes $H=1$, regardless of the other five signed values. Under a different selector assignment, another branch determines the answer. This family targets nested computation and consistent reuse of facts. It also illustrates why essentiality over all assignments does not imply that all ten facts are needed for every particular image: inactive branches can be irrelevant to the observed answer.

\subsubsection{Dense conjunctive normal form}
\label{app:family-dense-cnf}

\textbf{Meaning and construction.}
A CNF is an AND of clauses, each of which is an OR of literals. The dense family samples $24$ distinct signed clauses, each using five distinct variables chosen from the ten inputs:
\[
 C_j=L_{j1}\lor\cdots\lor L_{j5},
 \qquad H=\bigwedge_{j=1}^{24}C_j,
\]
where each $L_{jk}$ is a selected variable or its independently sampled negation. Duplicate signed clauses are rejected; clauses on the same five variables with different signs remain distinct. A proposal is also rejected unless every pair of variables co-occurs in at least one clause. Consequently, the primal graph, whose vertices are variables and whose edges indicate clause co-occurrence, is the complete graph $K_{10}$. In addition to the shared truth-table filters, the generator requires a certified lower bound of at least five on the clause--variable incidence graph's treewidth. This bipartite graph connects each clause to its variables and is distinct from the primal graph. The released incidence bounds are reported separately in this appendix.

\textbf{Worked evaluation and diagnostic role.}
For example, a sampled clause might be
$C_1=x_1\lor\neg x_2\lor x_3\lor x_4\lor\neg x_5$.
It is false at $(x_1,x_2,x_3,x_4,x_5)=(0,1,0,0,1)$, making the whole CNF false regardless of its other clauses. If it is true, the remaining clauses must still all hold. Variables recur across clauses, possibly with opposite signs, so satisfying one local requirement does not establish the complete answer. This family targets structural coupling through many overlapping constraints, and the graph checks certify this structure.

\subsubsection{Sparse chain CNF}
\label{app:family-sparse-cnf}

\textbf{Meaning and construction.}
The sparse control orders the ten variables by the sampled permutation, writing $u_i=x_{\pi(i)}$, and constructs eight clauses on successive triples:
\[
 H=\bigwedge_{k=1}^{8}
 \bigl(L_{k,k}\lor L_{k,k+1}\lor L_{k,k+2}\bigr),
 \qquad L_{k,j}\in\{u_j,\neg u_j\}.
\]
Each occurrence's sign is sampled independently, including when the same variable recurs in neighboring clauses. Adjacent clauses share two variables, clauses two positions apart share one, and more distant clauses share none. This fixed pattern produces a chain of overlapping triangles in the primal graph, of treewidth two. The common output-fraction and influence filters are then applied; unlike the dense family, there is no requirement to connect every pair of variables.

\textbf{Worked evaluation and diagnostic role.}
Two adjacent clauses could be
$(u_1\lor u_2\lor\neg u_3)$ and
$(\neg u_2\lor u_3\lor u_4)$.
At $(u_1,u_2,u_3,u_4)=(0,0,1,1)$, the first is false and the second is true, so the complete conjunction is false. Figure~\ref{fig:release-synthetic} gives a full released example of the same mechanism. This family is a control for more local constraint interactions since any false width-three CNF has a certificate of size at most three, given by a falsified clause. Comparing it with dense CNF varies clause count, clause width, and connectivity together.

\subsubsection{Polarity-trap composition}
\label{app:family-polarity-trap}

\textbf{Meaning and construction.}
This family embeds a CNF subformula that contradicts a literal-frequency heuristic, then combines its value with parity. With the same signed-literal convention as above, define
\[
\begin{aligned}
 D={}&(\neg\ell_1\lor\ell_2)
       \land(\neg\ell_1\lor\neg\ell_2)
       \land\bigwedge_{j=3}^{5}(\ell_1\lor\ell_j),\\
 P={}&\ell_2\oplus\ell_6\oplus\ell_7\oplus\ell_8
       \oplus\ell_9\oplus\ell_{10},
 \qquad H=D\oplus P.
\end{aligned}
\]
In $D$, the signed quantity $\ell_1$ occurs unnegated three times and negated twice. A frequency rule would recommend $\ell_1=1$. Yet that choice reduces the first two clauses to $\ell_2\land\neg\ell_2$, which is impossible. Satisfying $D$ instead requires $\ell_1=0$ and $\ell_3=\ell_4=\ell_5=1$, leaving $\ell_2$ free. Its inclusion in $P$ ensures that it still affects the full function. The generator normalizes signed literals to the underlying variables, counts their positive and negative occurrences, and exhaustively verifies that the core is satisfiable but cannot be satisfied with the majority-polarity recommendation for a certified trap variable.

\textbf{Worked evaluation and diagnostic role.}
At signed values $(0,0,1,1,1,0,0,0,0,0)$, the core has value $D=1$ and the parity has value $P=0$, hence $H=1$. Changing only $\ell_1$ to one makes $D=0$ and $H=0$, despite increasing agreement with its occurrence majority. This is a certified example of a misleading local cue for satisfying the CNF core. Appendix~\ref{sec:cnf-criteria} formalizes the certification and its scope.

\subsubsection{Ten-input parity control}
\label{app:family-parity}

\textbf{Meaning and construction.}
The final family applies parity directly to the ten signed inputs:
\[
 H=\ell_1\oplus\ell_2\oplus\cdots\oplus\ell_{10}.
\]
Variable permutation and random signs vary the displayed expression, which is expanded recursively into AND/OR/NOT. With all signs positive, three true statements give $H=1$; changing one further statement from false to true gives $H=0$. With negative signs, the count is over the signed literal values rather than the unnegated visual statements. Signs and the optional output negation yield only parity or its complement as Boolean functions, although their displayed trees and statement mappings vary.

\textbf{Diagnostic role.}
Flipping any one input changes the output. Consequently, each input has influence one and every assignment has certificate size ten over the unrestricted Boolean domain. However, the function admits a simple odd/even counting procedure. It therefore controls for broad dependence without assuming that needing many facts necessarily entails complex composition. Unlike parity of conjunctions or majorities, it requires no intermediate grouping or threshold decision. Appendix~\ref{app:compatibility} discusses these properties and the distinction between function complexity and its chosen representation.

\textbf{Relation to the broader design criteria.}
The families emphasize complementary operations rather than assigning one isolated difficulty factor to each group. Dependence on multiple facts and predictability from small subsets are measured across families; the polarity trap provides a specific heuristic counterexample; dense and sparse CNFs contrast structural coupling; and multiplexers make conditional nesting explicit. Every family receives the same equivalent-expression pairing procedure, so semantic invariance is tested across the full collection rather than by a separate family.  These constructions specify controlled logical properties; whether they challenge a particular model is an empirical question.

\subsection{Exact function and structure measurements}

For each pair of inputs, we consider the most accurate predictor that observes only those inputs. The pair is fixed across assignments, but its prediction may depend on their observed values. Under the uniform distribution on the full Boolean domain, the best achievable accuracy is
\[
  A_2(F)=
  \max_{\substack{T\subseteq\{1,\ldots,10\}\\|T|=2}}
  \frac{1}{2^{10}}
  \sum_{b\in\{0,1\}^{T}}
  \max_{y\in\{0,1\}}
  \left|\{x\in\{0,1\}^{10}:x_T=b,\ F(x)=y\}\right|.
\]
For each of the four observed value combinations, the optimal predictor returns the more frequent output among compatible assignments. We assign the weak-local-clues tag when $A_2(F)\le0.60$. Unlike this approximate prediction criterion, $c_F(a)$ requires exact determination of the answer and may use a different sufficient subset for each depicted assignment $a$.

We measure the displayed AST's leaf occurrences, total nodes, edge depth, repeated variables, negation nesting, and AND/OR alternation along root-to-leaf paths. For alternation counts, we ignore NOT nodes and collapse consecutive occurrences of the same binary operator. These are properties of a representation: rebracketing can change depth without changing the Boolean function. Given the input values, evaluating an AST takes time linear in its size.

The 6,000 displayed expressions contain 54--370 nodes (median 254) and have edge depth 6--15 (median 11). Among the 3,000 base expressions, 2,746 truth tables are distinct under fixed variable order. Of the 6,000 question records, 4,166 meet the broad-dependence criterion, 3,030 meet the weak-local-clues criterion, and 3,656 have at least five AND/OR alternations on some path. Certificate size is three for 330 records and ten for 520 records.

For the CNF constructions, we also measure incidence treewidth. The incidence graph has one vertex for each variable and one for each clause, with an edge whenever a variable occurs in a clause. The 1,000 dense-CNF questions have verified treewidth bounds of 8--9, including exact width nine for 830 questions. The sparse-CNF core has width two.

\subsection{What the polarity-trap annotation certifies}
\label{sec:cnf-criteria}
\label{sec:decoys}
Let $C$ be a satisfiable CNF core, and let $p_i$ and $q_i$ count positive and negative occurrences of variable $x_i$. For variables without a tie, the occurrence-polarity heuristic recommends $h_i=1$ when $p_i>q_i$ and $h_i=0$ when $p_i<q_i$. Define the certified trap set
\begin{equation}
  T_h(C)=\{i:p_i\ne q_i,\;
  C\land(x_i=h_i)\text{ is unsatisfiable}\}.
\end{equation}
Every assignment satisfying $C$ must set $x_i=1-h_i$ for each $i\in T_h(C)$.

For example, let $r\ge3$ and consider
\[
  C=(\neg x\lor z)\land(\neg x\lor\neg z)
  \land\bigwedge_{j=1}^{r}(x\lor u_j).
\]
The first two clauses force $x=0$, and the remaining clauses then force every $u_j=1$. However, $x$ occurs positively $r$ times and negatively twice, so the heuristic recommends $x=1$. This supplies a counterexample even within 2-CNF. Hob-VL records certified cores in 500 questions, corresponding to 250 base--rewrite pairs.

\subsection{Certificates, parity, and realizable assignments}
\label{app:compatibility}
\label{sec:feasible-domain}

If a CNF has maximum clause width $k$ and $F(a)=0$, fixing the values in any clause falsified by $a$ certifies the answer. Thus $c_F(a)\le k$, including when completions are restricted to a feasible subdomain containing $a$. In particular, a false width-three CNF has certificate size at most three and cannot satisfy the threshold $c_F(a)\ge5$. Such instances therefore provide a comparison group rather than meeting the broad-dependence criterion.

Parity has influence one for every input and certificate size ten on every assignment in the full ten-bit domain. For a direct CNF of $n$-bit parity without auxiliary variables, every non-tautological implied clause must contain all $n$ variables: otherwise flipping an omitted bit would produce both parities among assignments falsifying it. Each clause excludes one assignment, so excluding all $2^{n-1}$ assignments of the wrong parity requires $2^{n-1}$ clauses. One clause per excluded assignment attains the bound. Auxiliary-variable encodings are a different representation model \citep{emdin2022parity}.

Visual relations can restrict the set of realizable truth assignments. Let $\mathcal{S}$ denote the admissible scene annotations for predicates $g$, and define
\[
  \Omega_g=\{a(S):S\in\mathcal{S}\}.
\]
For $a\in\Omega_g$, let $c_{F,\Omega_g}(a)$ be the certificate size obtained by restricting the completions $z$ in Equation~2 to $\Omega_g$. Because this restriction removes possible completions,
\[
  c_{F,\Omega_g}(a)\le c_F(a).
\]
The generated-scene predicates admit the full Boolean domain. For photographs, the reported influence, certificate, and prediction measurements use the unrestricted abstract domain. They characterize the Boolean function without asserting that every assignment or single-bit change is physically realizable. The reference answer at the recorded assignment is unaffected by this distinction.

\subsection{Image selection and identification coverage}
\label{app:image-review}
Starting from 55 source images, we excluded seven duplicate photographs, one game screenshot, and one image with an existing screen frame. The retained 46 photographs belong to 45 scene families. Each photograph supports 42 or 44 Yes/No questions and 21 or 22 identification questions. The 1,000 identification questions use nine templates in approximately equal proportions and cover 241 distinct photograph--target combinations. The target is also named as a reference object in 541 questions. Among the 7,524 candidate evaluations, 1,789 include an unknown comparison; there are 1,968 unknown candidate--predicate comparisons in total. Each question has the same unique target under every completion of these unknown values. Under the finite-candidate identification criterion, the minimum number of predicate values needed to identify the target is two for 870 questions and three for 130 questions.

For generated images, automated checks compare color, shape probes, label ink, object bounds, and object separation against the renderer specifications. Eight generated images also received visual inspection by an AI assistant. Photographic facts and approximate object reference points include AI-assistant annotations. In addition, these tasks also underwent a rigorous human audit.

\section{Prompt Grammar and Released Examples}
\label{app:language}

For Yes/No questions, the symbolic prompt lists the visual statements and then
gives a fully parenthesized Boolean expression. The structured natural-language
(NL) compiler traverses the same abstract syntax tree (AST), replacing each
variable occurrence with its corresponding statement. It renders AND as
``Both of the following conditions hold,'' OR as ``At least one of the following
conditions holds, possibly both,'' and NOT as ``It is not the case that the
following condition holds.'' Indentation identifies each operator's immediate
subconditions. Thus, OR remains inclusive, NOT applies to its entire child, and
repeated variables retain the same visual referents. 

Symbolic Yes/No prompts contain 182--535 whitespace-delimited words (median: 371), whereas NL prompts contain 524--3,611 words (median: 1,911.5). These are word counts, not model-specific token counts. Then, a separate parser reconstructs the AST of each Yes/No NL prompt exactly. Identification prompts use a separate template compiler and are evaluated over the listed candidates. Their validation checks regenerated prompts and candidate answers, but does not independently parse the NL text back into an AST. The examples we provide below preserve the released wording, and all omissions are marked explicitly.

\subsection{Released examples in both formats}
\label{app:release-examples}
 
\begin{figure}[H]
\centering
\begin{minipage}[t]{0.40\linewidth}
\vspace{0pt}
\includegraphics[width=\linewidth]{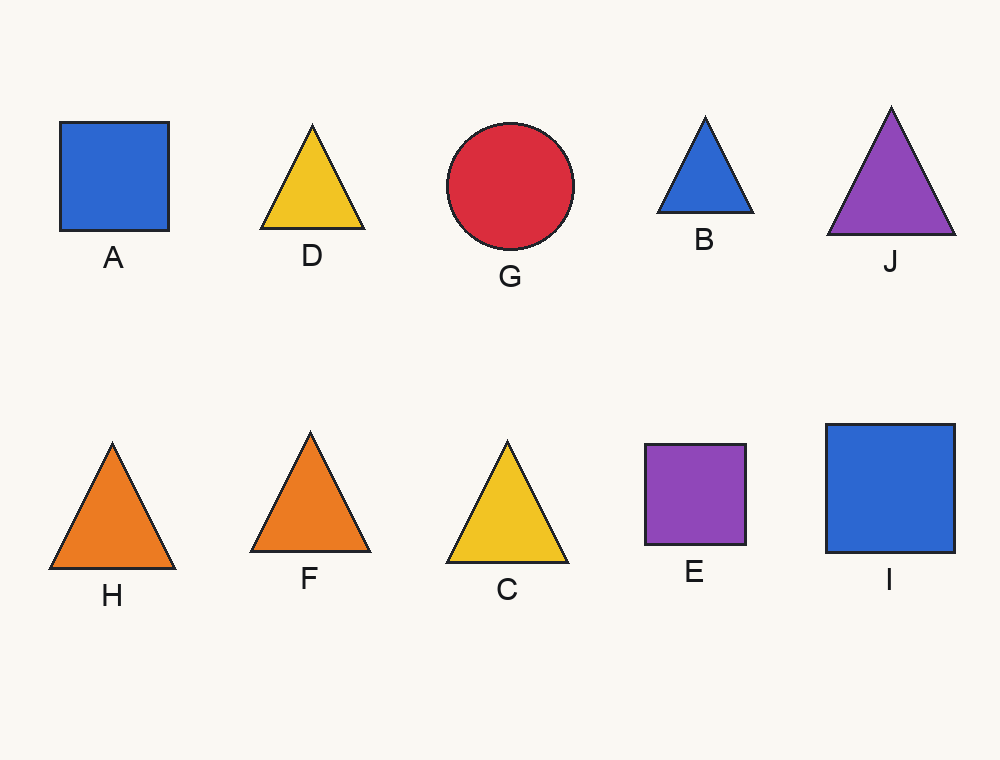}
\end{minipage}\hfill
\begin{minipage}[t]{0.57\linewidth}
\vspace{0pt}
\begin{lstlisting}[style=prompt]
Evaluate the condition using the labeled image. Each p-number means the visual statement listed below. AND requires both operands; OR is inclusive (one or both); NOT reverses the truth value. Parentheses specify all grouping. [...] A label names only its designated object. Answer Yes if the entire condition is true; otherwise answer No.

Statements:
p1: Object C is yellow.
p2: Object B is a triangle.
p3: Object H is a triangle.
p4: Object A is blue.
p5: Object I is orange.
p6: Object D is orange.
p7: Object G is a triangle.
p8: Object J is a circle.
p9: Object F is a triangle.
p10: Object E is green.

Condition:
((((p2 OR ((NOT p3) OR p4)) AND ((NOT p3) OR (p4 OR (NOT p10)))) AND ((p1 OR (p4 OR p10)) AND (p1 OR (p9 OR p10)))) AND ((((NOT p1) OR (p7 OR p9)) AND (p7 OR ((NOT p8) OR (NOT p9)))) AND ((p6 OR (p7 OR p8)) AND (p5 OR ((NOT p6) OR p8)))))
\end{lstlisting}
\end{minipage}
\caption{A symbolic Yes/No example from the sparse-CNF control family. The prompt
is reproduced verbatim except for one instruction sentence about spatial
conventions, marked \texttt{[...]}. The image is \texttt{synthetic\_0362}, and the
expression contains 54 AST nodes, the smallest expression size in the release.
The answer is \emph{No}: D is yellow, G is a circle, and J is a triangle, so
$p_6=p_7=p_8=0$ and the clause $p_6\lor p_7\lor p_8$ is false. These three values
certify the answer. A false width-three CNF always has a certificate of size at
most three  (Appendix~\ref{app:compatibility}), so this control does not meet the benchmark's
broad-dependence criterion. Its De Morgan partner rewrites two subtrees and
contains 60 AST nodes; one rewrite replaces
$(p_6\lor(p_7\lor p_8))\land R$ with
$\neg\bigl(\neg(p_6\lor(p_7\lor p_8))\lor\neg R\bigr)$,
where $R$ denotes the other child of that conjunction.}
\label{fig:release-synthetic}
\end{figure}

\begin{figure}[t]
\centering
\begin{minipage}[t]{0.36\linewidth}
\vspace{0pt}
\includegraphics[width=\linewidth]{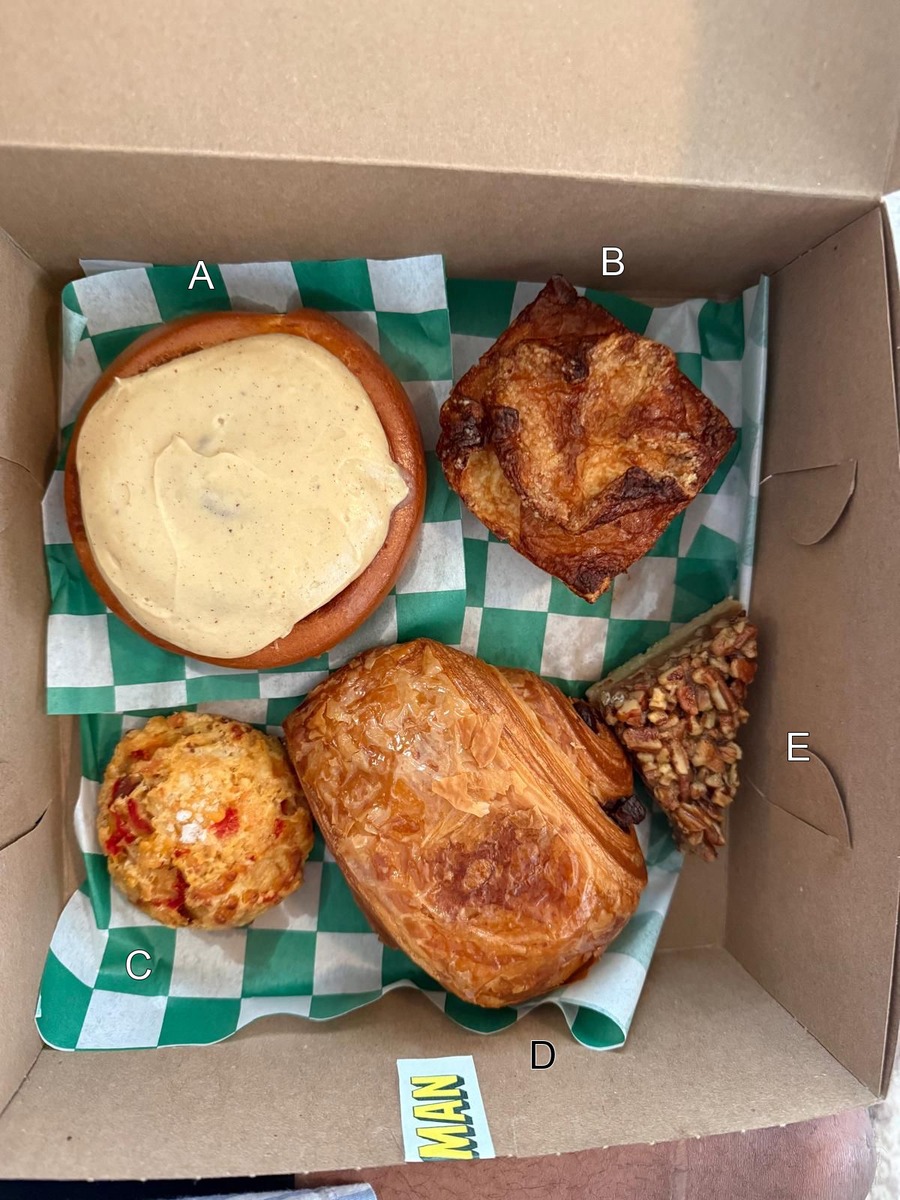}
\end{minipage}\hfill
\begin{minipage}[t]{0.61\linewidth}
\vspace{0pt}
\begin{lstlisting}[style=prompt]
Find the single labeled object that satisfies the entire description. Consider every label listed below, including objects used as references. Directions refer to the centers of the designated objects in image coordinates, not the positions of the label letters or physical depth. An object is neither left of, right of, above, nor below itself. Answer with only its label.

Candidate labels: A, B, C, D, E

Description (indentation groups subconditions):
- Both of the following conditions hold:
 - At least one of the following conditions holds, possibly both:
  - The candidate object is to the right of object B.
  - It is not the case that the following condition holds:
   - The candidate object is above object C.
 - At least one of the following conditions holds, possibly both:
  - The candidate object is to the left of object D.
  - The candidate object is above object E.
\end{lstlisting}
\begin{lstlisting}[style=prompt]
[same header and candidate labels]
p1: The candidate object is to the right of object B.
p2: The candidate object is above object C.
p3: The candidate object is to the left of object D.
p4: The candidate object is above object E.

Condition:
((p1 OR (NOT p2)) AND (p3 OR p4))
\end{lstlisting}
\end{minipage}
\caption{An identification question for \texttt{photo\_001}. The upper block
shows the complete NL prompt; the lower block shows the symbolic statements and
condition, with the shared header and candidate list abbreviated. The correct
label is C: $p_2$ is false because C is not above itself, and $p_3$ is true because
C lies left of D. Thus, both disjunctions in
$(p_1\lor\neg p_2)\land(p_3\lor p_4)$ hold. D's vertical relation to C falls
within the 6\% uncertainty margin and is recorded as unknown, but D is excluded
because it is neither left of itself nor above E, making both $p_3$ and $p_4$
false. The values of $p_2$ and $p_3$ distinguish C from every other candidate.}
\label{fig:release-identification}
\end{figure}

The following examples are drawn from Hob-VL.
Figure~\ref{fig:release-synthetic} shows a symbolic Yes/No prompt, Figure~\ref{fig:release-language} its natural-language counterpart, and Figure~\ref{fig:release-identification} an
identification question in both formats. Marked omissions shorten the displayed prompts. Reference answers and explanatory captions are provided only for the reader and were not included in model inputs.

\begin{figure}[t]
\centering
\begin{lstlisting}[style=prompt]
Evaluate the condition using the labeled image. Each bullet states a condition; its indented bullets are its immediate subconditions. Both requires the two immediate subconditions to hold. At least one allows either or both immediate subconditions to hold. It is not the case reverses the truth of the entire immediate subcondition. [...] Answer Yes if the entire condition is true; otherwise answer No.

Condition:
- Both of the following conditions hold:
 - Both of the following conditions hold:
  - Both of the following conditions hold:
   - At least one of the following conditions holds, possibly both:
    - Object B is a triangle.
    - At least one of the following conditions holds, possibly both:
     - It is not the case that the following condition holds:
      - Object H is a triangle.
     - Object A is blue.
   - At least one of the following conditions holds, possibly both:
    - It is not the case that the following condition holds:
     - Object H is a triangle.
    - At least one of the following conditions holds, possibly both:
     - Object A is blue.
     - It is not the case that the following condition holds:
      - Object E is green.
  - Both of the following conditions hold:
   - At least one of the following conditions holds, possibly both:
    - Object C is yellow.
    - At least one of the following conditions holds, possibly both:
     - Object A is blue.
     - Object E is green.
   [... one further clause of this block, 5 lines ...]
 - Both of the following conditions hold:
  [... the four remaining clauses, 26 lines ...]
\end{lstlisting}
\caption{The natural-language format of the item in
Figure~\ref{fig:release-synthetic}, sharing its image, answer, and
identifier. The opening lines are verbatim; the two bracketed elisions
replace 31 of the 54 condition lines so the figure fits on the page.
The complete prompt has 524 words, versus 183 for the symbolic version,
because the statement for each of the 24 leaf occurrences is written out in
place. Indentation is part of the input and is preserved in the released
JSONL.}
\label{fig:release-language}
\end{figure}

\clearpage

\section{Executable Validation and Provenance}
\label{app:checks}

The independent \texttt{scripts/dataset/validate\_dataset.py} parser and NumPy evaluator check the Boolean expressions, all $1{,}024$ assignments, influences, certificates, two-statement predictors, rewrite traces, polarity-trap cores, and graph witnesses without importing the generator. The separate \texttt{validate\_rendering.py} probes PNGs against scene descriptions. The stored reports include counts, checksums, and explicit validation limits.

From the repository root, the following commands validate the NL and
identification collections and regenerate the result tables:
\begin{verbatim}
python scripts/dataset/build_natural_language.py --check
python scripts/dataset/build_photo_identification.py \
    --count 1000 --check
python -m evals.paper_results
\end{verbatim}

The identification command requires \texttt{---count 1000}; omitting this argument
checks the earlier 46-example pilot. Its checks reuse generation functions and
verify that exactly one candidate satisfies each description under every
completion of unknown values. The exhaustive examples in
\texttt{scripts/check\_examples.py} additionally check the illustrative
certificate and parity arguments, and we augment them with a rigorous human audit.\looseness=-1

Evaluation histories retain raw responses, parsing outcomes, finish reasons,
token usage, and retry records. Before generating a result table, the reporting
code checks that every planned request has a final response after retries, verifies the
reference answers and evidence shared by paired prompts, and compares saved
summary counts with counts recomputed from the histories. Models receive only
the prompt and prepared image; reference answers and annotation records are used
locally for validation and scoring. Reconstructing the dataset requires the source images, Python~3.11 or later,
Pillow, NumPy, and the font configured by the renderer. Model aliases and provider
defaults can change, limiting exact reproduction of API outputs. 

\end{document}